\documentclass[11pt]{article}         
\usepackage[margin=1in]{geometry} 
\usepackage{fullpage}     
\usepackage[utf8]{inputenc} 
\usepackage[T1]{fontenc}    
\usepackage{url}            
\usepackage{booktabs}       
\usepackage{amsfonts}       
\usepackage{nicefrac}       
\usepackage{microtype}      
\usepackage{bm}             
\usepackage[colorlinks=true, citecolor=blue]{hyperref}
\usepackage{natbib}
\usepackage{algorithm}
\usepackage{algorithmic}
\usepackage{enumitem}
\usepackage{graphicx}
\usepackage{amssymb}
\usepackage{xcolor}
\usepackage{dsfont} 
\usepackage{array}
\newcolumntype{P}[1]{>{\centering\arraybackslash}p{#1}}
\usepackage{amsmath}
\usepackage{setspace}

\usepackage{caption}
\usepackage[tableposition=above]{caption} 
\usepackage{subcaption}
\usepackage{placeins}
\usepackage[english]{babel}
\usepackage{amsthm}
\usepackage{tabularx,booktabs}

\def\code#1{\texttt{#1}}
\newcolumntype{Y}{>{\centering\arraybackslash}X}

\theoremstyle{definition}

\theoremstyle{definition}
\newtheorem{remark}{Remark}
\theoremstyle{definition}

\usepackage{tabularx,booktabs}
\newcolumntype{Y}{>{\centering\arraybackslash}X}

\usepackage{enumitem} 
\setlist{noitemsep} 

\usepackage{datetime}                       
\newdateformat{mydate}{\monthname[\THEMONTH] \THEYEAR}   

\title{Estimation of Large Financial Covariances: \\A Cross-Validation Approach\footnote{
The authors would like to thank Daniel Bartz, Zdzis\l aw Burda, Mihai Cucuringu, Lisa Goldberg, and Andrzej Jarosz as well as seminar participants at the University of Oxford and Man AHL for the helpful comments. The authors are also grateful to Christian Bongiorno for providing research assistance and valuable feedback. This work was supported by the Oxford-Man Institute of Quantitative Finance. 
}}

\definecolor{electricultramarine}{rgb}{0.25, 0.0, 1.0}   

\author{
Vincent Tan\thanks{Oxford-Man Institute, Department of Engineering Science. E-mail:  \href{mailto:vincent.tan@eng.ox.ac.uk}{vincent.tan@eng.ox.ac.uk}}
 \and Stefan Zohren\thanks{Oxford-Man Institute, Department of Engineering Science. E-mail: \href{mailto:stefan.zohren@eng.ox.ac.uk}{stefan.zohren@eng.ox.ac.uk}}}

\begin{document}
\begin{titlepage}

\newcommand*{\thisdraft}{This version: January 2023} 
\newcommand*{\firstdraft}{First version: December 2020}  

\date{\firstdraft \\ \thisdraft}
\maketitle
\thispagestyle{empty}

\setstretch{1.15}
\begin{abstract}

We introduce a novel covariance estimator for portfolio selection that adapts to the non-stationary or persistent heteroskedastic environments of financial time series by employing exponentially weighted averages and nonlinearly shrinking the sample eigenvalues through cross-validation. Our estimator is structure agnostic, transparent, and computationally feasible in large dimensions. By correcting the biases in the sample eigenvalues and aligning our estimator to more recent risk, we demonstrate that our estimator performs well in large dimensions against existing state-of-the-art static and dynamic covariance shrinkage estimators through simulations and with an empirical application in active portfolio management.

\end{abstract} \medskip

\begin{flushleft} JEL Classification: C13, C58, G11. \\ \medskip Keywords: Cross-validation, exponentially weighted averages, large-dimensional statistics, nonlinear shrinkage, portfolio selection, rotation equivariance. \end{flushleft}

\end{titlepage}

\clearpage

\setstretch{1.15}
\section{Introduction}

The covariance matrix of asset returns is a central object in the portfolio selection problem of \cite{markowitz}. In practical applications, the covariance matrix is not known \textit{a priori} and we have to resort to estimators. When estimating a large covariance matrix, one confronts the `curse of dimensionality', where the problem size grows quadratically as the number of assets increases. This means that we need to have a long enough historical data set to produce reliable estimates of the covariances. For example, to reliably estimate a covariance matrix of $1000$ assets, at least $40$ years of daily data is needed. The estimation of large numbers of risk parameters relative to the number of observations introduces the possibility of aggregating many estimation errors into a global significant error. When such a matrix is used in the context of portfolio optimization,  \cite{jobson1980estimation} and \cite{chopra1993effect}, among others, showed that such optimizers exhibit poor out-of-sample properties, and inevitably become what \cite{michaud} regards as ``error maximization'' procedures.


To improve the estimation of a covariance matrix, it is thus imperative to reduce the effective number of risk parameters that require estimation. In finance, factor models are often employed for this purpose with early developments stemming from the single factor model of \cite{sharpe1963simplified} and its multi-factor extension in the context of the Arbitrage Pricing Theory (APT) introduced by \cite{ross1976arbitrage}, and later generalized by \cite{RePEc:ecm:emetrp:v:51:y:1983:i:5:p:1281-304}. Some examples that reify the APT include the fundamental risk factor model of \cite{fama1992cross} and commercial risk models that have been built and maintained by companies such as BARRA, Northfield, Axioma, and so on. What is common in these approaches is that the underlying structure of the covariance is being modeled. As a result, the estimation error is typically reduced but at the expense of increased model bias.



Another line of attack is the shrinkage principle introduced by \cite{stein1956inadmissibility} and \cite{james1961estimation} in the context of the estimation of the means. Shrinkage can be understood as a form of \textit{regularization} -- to use machine learning parlance -- to control the effects of sampling variation, which results in overfitting. This idea was extended to covariance matrices by \cite{haff1980empirical} and  \cite{ledoit2004well} by taking a linear combination of a sample covariance matrix and a (scaled) identity matrix in order to produce a composite matrix that has a favorable model bias-variance trade-off. Besides the identity matrix, other types of structured target matrices considered in the literature include the constant correlation model \citep{ledoit2004honey}, or the single factor model \citep{ledoit2003improved}. However, a limitation of the linear shrinkage estimator is that it is constrained by a single common shrinkage intensity, and hence lacks the capacity to suppress further variations in the data. 

A more recent and powerful approach that extends upon the linear shrinkage technology attempts to reduce the estimation error without assuming any \textit{a priori} structure on the covariance. This is achieved by shrinking the eigenvalues at distinct intensities. The inception of this idea goes back to \cite{stein1, stein2} and there are two modern approaches on this front. The first is an analytical approach taken by \cite{ledoit2011eigenvectors} and subsequently, \cite{bbp}, where the authors utilized the tools from Random Matrix Theory in order to provide a theoretical formula for shrinking the eigenvalues. There are two possible approaches to implementing their formula: discretization \citep{ledoit2015spectrum} or kernel estimation \citep{ledoit2020analytical, ledoit2022quadratic}. In parallel to this development, there is also a numerical scheme from \cite{abadir2014design}, \cite{lam2016nonparametric} and \cite{bartz}, where the authors demonstrate that we can accomplish a similar shrinkage effect by means of sample splitting and cross-validation. 

Most of the aforementioned developments, however, revolve around a uniformly weighted covariance matrix that weighs every observation equally. However, in finance, it is known since \cite{mandelbrot1963variation} that financial returns exhibit certain empirical regularities such as volatility clustering. This volatility clustering effect can overwhelm our observations, which leaves us with effectively fewer observations than what we originally had because the stock returns are dominated by days with large fluctuations. Thus, the use of exponentially weighted average schemes such as RiskMetrics developed by \cite{rm96} and \cite{zumbach} is arguably more relevant for non-stationary or heteroskedastic environments since it restricts the effective sample size so that more local observations are used. However, these models are also susceptible to estimation errors in large dimensions since the effective time scale used to measure the observations is smaller than the sample size.  



In this paper, we introduce a simple yet effective covariance estimator that is tailored specifically to address the curse of dimensionality present in exponentially weighted average covariances. In particular, we extend the cross-validation shrinkage procedure of \cite{abadir2014design}, \cite{lam2016nonparametric} and \cite{bartz}, towards exponentially weighted average covariances in order to robustify them against large dimensions. The broad appeal of cross-validation is that it is a well-known method that many researchers and practitioners are familiar with, and it is an effective procedure to combat overfitting. The key idea behind our method is to split the original observations into two independent sets; the observations from the `test' set are used to correct the biases that are embedded in the estimated eigenvectors from the `training' set due to in-sample overfitting. As a result, extreme eigenvalues are pulled towards more central values to compensate for those biases. Our method is computationally scalable to large dimensions and it demonstrates superior results in this regime through an empirical application of active portfolio management. 



The paper is structured as follows. Section \ref{sec:background} introduces the exponentially weighted average covariance matrix and its decomposition. Section \ref{sec:proposed} describes our proposed cross-validation-based nonlinear shrinkage covariance estimator. Section \ref{sec:montecarlo} contains the simulation study. Section \ref{sec:empiricalchp1} provides the empirical experiments. Section \ref{sec:conclusionchp1} concludes. Appendices \ref{appendix:figureschp1}--\ref{sec:mps} contain all the figures, tables, and experimental results for the Markowitz portfolio with signal problem.

\section{Setting the Stage}\label{sec:background}

\subsection{The Weighted Sample Covariance Matrix}

Let $X := (x_1, \ldots , x_T)'$ be a $T \times N$ matrix of $T$ observations of a vector of assets, $x_t$, each of dimensionality $N$. Here, the symbol $(')$ denotes the transpose operator of a vector or matrix, the symbol $:=$ is a definition sign, and the symbol $=$ is the equal sign. The vector of assets is assumed to have zero means and a population covariance matrix $\Sigma$, which is unobserved to the investor. The population covariance in our study may potentially be time-varying over the sample period, in which case, we shall assign a time-index subscript to it, for example, $\Sigma _t$ for $t=1, \ldots, T$.  

The classical estimator for $\Sigma$ is the sample covariance matrix is defined as
\begin{align}\label{eq:SC}
    S :=  \frac{1}{T} X'X .
\end{align}
The standard sample covariance is an estimator of the population covariance since it is unbiased (that is, its expectation coincides with the population covariance matrix). In general, a standard sample covariance can be generalized to include some arbitrary weight profile assigned along the time dimension. In particular, it is expressed in the following form
\begin{align}\label{eq:weightedcov}
    S_W := \frac{1}{T} X' W X, 
\end{align}
where $W$ is a fixed symmetric and positive-definite $T \times T$ matrix. This weighting scheme generalizes the standard sample covariance and it provides us a way to incorporate prior knowledge about the non-independence and non-identicality nature of the observations in the covariance matrix. Indeed, if $W = \mathbb{I}_T$, where $\mathbb{I}_T$ is an $T$-dimensional identity matrix, then we recover the expression for the standard sample covariance in \eqref{eq:SC}.  

Without loss of generality, we assume that $T^{-1} \mathsf{Tr}(W) = 1$ where $\mathsf{Tr}(\cdot)$ denotes the trace of a square matrix or the sum of its diagonal entries. This ensures that these weighted covariance matrices are also unbiased estimators, that is $\mathbb{E}[S_W] = \Sigma $. In other words, the weighting of observations in time should not affect the recovery of the ground truth on average. 

\subsection{Exponentially Weighted Average Covariance Matrix}

To address the problem of non-stationarity in the time-series setting, we look at a (sample) covariance matrix that is measured using an exponentially weighted average scheme, henceforth, EWA-SC, which is given by
\begin{align}\label{eq:emaSC}
     E := \frac{1 - \beta}{1 - \beta ^T} \sum ^{T}_{t=1} \beta ^{T-t} x_{t} x_{t} '.
\end{align}
Here, $\beta \in (0, 1)$ is the exponential decay rate. The timescale for which the observations are ``effectively'' measured with this scheme can be determined using a characteristic time-scale defined as $\tilde{T} := 1/(1-\beta)$. For example, if $\beta = 0.999$, then the timescale would be $1000$ days (or roughly 4 years). In the limit where $\beta$ tends to one from below the EWA-SC converges towards the standard sample covariance; conversely, if $\beta$ moves away from one, it places more weight on the recent cross-products observations. 

The EWA-SC in Equation \eqref{eq:emaSC} has a weight matrix profile whose diagonal entries are
\begin{align}
    W_{t,t} = T \frac{1-\beta}{1 - \beta ^T} \beta ^{T-t} , 
\end{align}
for $t = 1, 2, \ldots , T$ and zero in the off-diagonal entries. Further, let us define the following auxiliary observation matrix:
\begin{align} \label{eq:auxiliary}
    \tilde{X} := W ^{1/2} X  .
\end{align}
Since the matrix $W$ is diagonal, its matrix factor $W ^{1/2}$ is simply the square root of its diagonal entries taken with a non-negative sign. Based on our redefinition of the observation matrix, we can see that the EWA-SC can be expressed in a similar form as the standard uniformly weighted sample covariance. The advantage of recasting the EWA-SC in this way is that many of the refinements for the standard sample covariance that have been developed over the years are at our disposal; this includes shrinkage.

\subsection{Decomposing the Covariance Matrix}

The analysis of a covariance matrix would be easier if the entries are uncorrelated. While this is not generally true for asset returns, it is nonetheless useful if we can summarize the contents of a matrix in terms of key low-dimensional objects with mutually uncorrelated components. Fortunately, the EWA-SC is symmetric and so for a fixed decay rate $\beta$, they admit the following spectral decomposition:
\begin{align}
    E = \sum ^N _{i=1} \hat{\lambda} _i \hat{u}_i  \hat{u}_i', \quad \text{and} \quad  \Sigma = \sum ^N _{i=1} \lambda _i u_i  u_i' ,
\end{align}
where $(\hat{\lambda} _1 ,\ldots , \hat{\lambda}_N; \hat{u}_1 , \ldots , \hat{u}_N)$ denotes a system of sample eigenvalues and eigenvectors of $E$, and $(\lambda _1 ,\ldots , \lambda_N; u_1 , \ldots , u_N)$ denotes a system of eigenvalues and eigenvectors of the population covariance $\Sigma$.\footnote{Both the sample eigenvalues and sample eigenvectors are implicitly parameterized by the decay rate $\beta$ since the underlying EWA-SC depends on it. We suppress their dependence on the decay rate for brevity.} The eigenvalues are assumed to be sorted in ascending order.

The main advantage of the sample eigenvalues and sample eigenvectors is that they are consistent estimators; that is, they recover the population eigenvalues and eigenvectors when the number of observations $T$ becomes large while the number of assets $N$ is held fixed. However, this desirable statistical property breaks down in financial applications where the number of assets is of the same order of magnitude in comparison to the number of observations. In particular, the sample eigenvalues and sample eigenvectors end up being biased estimates of their population counterparts. We defer to \cite{burda2011applying} for an in-depth explanation of this phenomenon for the EWA-SC matrix. 

\section{Shrinkage Estimation}\label{sec:proposed}

\subsection{Covariance for Portfolio Selection}

In light of the previous discussion, we consider the framework of \cite{stein1, stein2} as a guide to correct the biases of the EWA-SC. It suggests that the sample eigenvalues should be corrected while retaining the sample eigenvectors of the original matrix. This can be formulated mathematically as:
\begin{align}\label{eq:rie}
    \hat{\Sigma}  = \sum ^N _{i=1} \xi _i \hat{u}_i  \hat{u}_i',
\end{align}
%
where $\xi := ( \xi _i )_{i=1,\ldots, N}$ is an $N$-dimensional vector that we have to obtain. This framework serves as a reasonable starting point to address the systematic biases in the sample covariance. Indeed, absent any \textit{a priori} knowledge about the structure of the covariance matrix, the most natural guess that we have about the population eigenvectors is the sample eigenvectors that we observe. Hence, the sample eigenvectors are the only ones that we get to keep and we have to try to make the best out of this situation by coming up with a set of bias-corrected eigenvalues so that the covariance matrix as a whole is `closer' to the truth. 

The question is how we should choose the vector of bias-corrected eigenvalues $\xi$. We answer this question by choosing the one that minimizes a loss function between estimators of the form \eqref{eq:rie} and the population covariance matrix. To determine the optimal estimator within the postulated form of \eqref{eq:rie}, we consider a loss function between the population and the estimated covariance matrix based on the minimum variance loss function proposed by \citet[Definition 1]{engle2019large}, which is tailored specifically for the purpose of portfolio selection:
\begin{align}\label{eq:loss}
    \mathcal{L}(\hat{\Sigma}, \Sigma ) := \frac{\mathsf{Tr}(\hat{\Sigma} ^{-1} \Sigma  \bm \hat{\Sigma} ^{-1}) / N}{(\mathsf{Tr}(\hat{\Sigma}^{-1}) / N)^2} - \frac{1}{\mathsf{Tr}( \Sigma ^{-1}) / N}.
\end{align}
The loss function can be interpreted as the excess out-of-sample portfolio variance that would be materialized from using an estimator instead of the population covariance matrix, and it should be reduced as much as possible. 


If we insert the postulated estimator \eqref{eq:rie} into Equation \eqref{eq:loss}, then from the cyclic property of the trace operator and orthogonality of the eigenvectors, we get
\begin{align}
    \mathcal{L}(\hat{\Sigma}, \Sigma ) = N \frac{\sum ^N _{i=1} \xi ^{-2}_i \hat{u}_i' \Sigma \hat{u}_i }{\left (\sum ^N _{i=1} \xi ^{-1}_i \right )^2} + \text{constant} .
\end{align}
By minimizing this loss function with respect to $\xi_i$, we have 
\begin{align}
    \xi _i ^\circ = \text{scalar} \times \hat{u}'_i \Sigma \hat{u}_i , \enskip \text{for } i=1,\ldots ,N, 
\end{align}
where the scalar value is independent of the asset index $i$. The scalar can be chosen such that the trace of the population covariance matrix is preserved. This can be achieved by setting the scalar to the value 1; see for example \citet[Proposition 4.1]{ledoit2017nonlinear}. 

We see that $\xi _i ^\circ$ is neither the sample eigenvalue nor the population eigenvalue. Rather, it is an `eigenvalue' that internalizes both the sample and population information in order to differentiate itself from its sample and population counterparts. From the economic perspective, one can interpret each $\xi _i ^\circ$ as an out-of-sample variance of a portfolio whose weights are the $i$th sample eigenvector $\hat{u}_i$. This is an enhancement over the sample eigenvalues $\hat{\lambda}_i$, which is an in-sample variance computed from a portfolio that uses the same observations as the weights $\hat{u}_i$. Further, \citet[Section 2.3]{ledoit2004well} showed that $\xi _i ^\circ$ is less dispersed cross-sectionally than the sample eigenvalue $\hat{\lambda}_i$. Hence, it constitutes a form of shrinkage itself that corrects the biases in the sample eigenvalues. 


If we plug $\xi^\circ$ into Equation \eqref{eq:rie}, we arrive at the following covariance estimator: 
\begin{align}\label{eq:fsopt}
    \hat{\Sigma}^\circ  = \sum ^N _{i=1} \xi _i ^\circ \hat{u}_i  \hat{u}_i', \enskip \text{where}  \enskip \xi ^\circ _i = \hat{u}_i ' \Sigma \hat{u}_i \enskip \text{for } i = 1, \ldots , N.
\end{align}
An advantage of $\xi _i ^\circ$ is that it is positive for $i = 1, \ldots , N$ as long as $\Sigma$ is positive definite. In this case, the covariance estimator \eqref{eq:fsopt} is positive definite and invertible when $N>T$. Unfortunately, $\hat{\Sigma}^\circ $ is not attainable in practice since it depends on the population covariance $\Sigma$, which is the very object that we wanted to estimate in the first place. While this seems paradoxical at first sight, it is possible to substantiate the estimator $\hat{\Sigma}^\circ $ with real data and this is what we aim to achieve with our proposed method in the next subsection.

\subsection{Out-of-Sample Variance Estimation by Cross-Validation} \label{sec:emacv}

Another perspective on the biased nature of the sample eigenvalues and sample eigenvectors is that they are objects that have been obtained through numerical optimization routines. Hence, they are likely to overfit a given data. To address this problem, cross-validation is a powerful scheme to mitigate overfitting and estimate the out-of-sample performance for independent and identically distributed (i.i.d.) observations. In our setting, the out-of-sample performance that we are interested in is $\xi^\circ_i$, which is the out-of-sample variance of the $i$th sample eigenvector portfolio for a pre-specified decay rate. 



%


To perform cross-validation, we start by splitting the $T$ (randomized) observations into $K$ non-overlapping index sets (or folds) given by $\{ \mathcal{I}_k | \mathcal{I}_k \subset \{ 1, \ldots , T \} \} ^K_{k=1}$. Each set indexed by $\mathcal{I}_k$ is called a `test' fold, whereas the set of indices of the remaining observations forms a `training' fold. From this construction, we can estimate the out-of-sample variance $\xi ^\circ _i$; we introduce our $K$-fold cross-validation estimator, which is defined as
\begin{align} \label{eq:proposed}
    \xi_i ^{\bullet} := \frac{1}{K} \sum ^K _{k=1} \sum _{t \in \mathcal{I}_k} \frac{1}{|\mathcal{I}_k|} \left (\hat{u}_i [k] ' \tilde{x}_t \right ) ^2, \enskip \text{for } i = 1, \ldots , N,
\end{align}
where $|\mathcal{I}_k|$ denotes the cardinality of the $k$th test set such that each of them is approximately equal in size, that is, $K|\mathcal{I}_k | \approx T$. Here, $\hat{u}_i [k]$ is the $i$th sample eigenvector of a sample covariance matrix that is obtained from the training fold, and $\tilde{x}_t$ is a sample vector of the auxiliary observation matrix in Equation \eqref{eq:auxiliary} from the test fold. This is a more general estimator of \cite{abadir2014design}, \cite{lam2016nonparametric} and \cite{bartz} since if $\beta$ tends to one, we recover the cross-validation estimator of the standard sample covariance as the auxiliary observation vectors become the original ones.

\begin{figure}[H] 
     \centering
     \caption{Schematic Overview of Five-Fold Cross-Validation}
    \captionsetup{font=footnotesize}
     \caption*{The figure displays a schematic diagram of a five-fold cross-validation scheme for estimating the vector of out-of-sample variances for the sample eigenvector portfolios. The light-shaded rectangles represent the training sets used to compute the sample eigenvectors, whereas the dark-shaded rectangles represent the testing sets used to estimate the vector of out-of-sample variances of the eigenvector portfolios.}
     \includegraphics[width=0.8\textwidth]{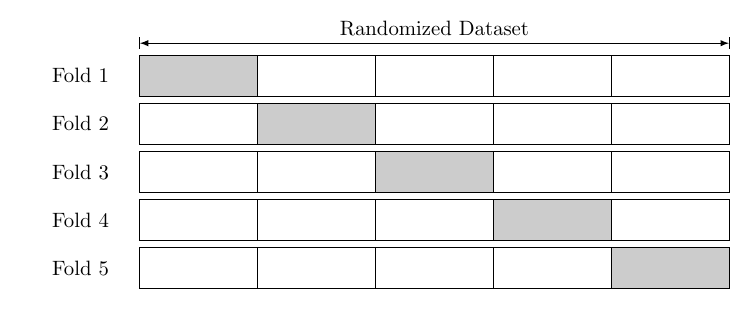}
    \label{fig:schematic}
\end{figure}

Figure \ref{fig:schematic} provides a schematic representation of the construction of the training and test sets for $K=5$ folds. We start by randomizing the auxiliary observation matrix $\tilde{X}$ from Equation \eqref{eq:auxiliary} along the time axis.\footnote{The randomization step in the algorithm is needed to break the temporal correlations of the auxiliary observation vector $\tilde{x}_t$ induced by the exponential weights so that the observations in each fold can be approximately independent. This allows the cross-validated eigenvalue reproduces the out-of-sample variance along the $i$th eigenvector $\hat{u}_i[k]$ through conditional averaging. Hence, they can be seen as an approximation to the optimal unattainable out-of-sample variance of the $i$th sample eigenvector portfolio $\xi ^\circ _i$.} We then split the (randomized) auxiliary observations into $K$ non-overlapping folds of equal size. Each of the $K$ folds is used as a test set, whereas all other folds are merged into a training set.  For each $K$ fold configuration, we estimate the sample eigenvectors from the training set and then estimate an $N$-dimensional vector of out-of-sample variances using the test set and the sample eigenvectors. Finally, we average the out-of-sample variance estimates over $K$ to give us the bias-corrected eigenvalue of the $i$th sample eigenvector portfolio denoted as $\xi^{\bullet}_i$ for all $i$. 

Finally, we plug $\xi^{\bullet}_i$ into the estimator \eqref{eq:rie} to yield our operational covariance estimator, which we term EWA-CV:
\begin{align}\label{eq:reconstr}
    \hat{\Sigma}  = \sum ^N _{i=1} \xi^{\bullet}_i \hat{u}_i  \hat{u}_i'.
\end{align}
By invoking a result in \citet[Corollary 4]{lam2016nonparametric}, EWA-CV readily yields a positive-definite covariance matrix as long as the population covariance is also positive-definite. An estimate for the inverse covariance can be obtained by taking the inverse of the covariance estimator. 

Note that our covariance estimator \eqref{eq:reconstr} is not entirely restricted to weight profiles that are exponentially declining in time. In principle, our auxiliary variable  definition \eqref{eq:auxiliary} can accommodate any weight matrix design $W$, provided it can be factorized into a matrix and a conjugate transpose. Thus, our estimator is general enough to incorporate other forms of weighting schemes to model the time dependence of asset returns or to be employed in other financial applications. We leave this investigation for future research.

\begin{remark}[Unbiasedness]
    The cross-validated eigenvalue in Equation \eqref{eq:proposed} is an unbiased estimator of the variance along the eigenvector directions $\hat{u}_i[k]$ conditional on the EWA-SC. Indeed, since the weighted covariances \eqref{eq:weightedcov} are unbiased estimators of the population covariance, then provided $\hat{u}_i [k]$ and $\tilde{x}_t$ are independent we have
\begin{align}
    \mathbb{E} [ \xi_i ^{\bullet} | E ]  =
    \hat{u}_i [k] ' \Sigma \hat{u}_i [k] .
\end{align}
In other words, the cross-validated eigenvalue reproduces the out-of-sample variance along the $i$th eigenvector $\hat{u}_i[k]$ through conditional averaging. Thus, they can be seen as an approximation to the unattainable out-of-sample variance of the $i$th sample eigenvector portfolio $\xi ^\circ _i$ from Equation \eqref{eq:fsopt}.
\end{remark}

\begin{remark}[Monotonicity of Eigenvalues]
    The ordering of cross-validated eigenvalues is often violated in practice. This may be unsettling for some researchers since it can be argued that the eigenvalues should satisfy a ranking where there they are ordered from the smallest to the largest; for example, see \cite{sheena1992inadmissibility}. To address this concern, we adopt the approach from \cite{stein1} and apply an isotonic regression of the set cross-validated eigenvalues on the sample eigenvalues of $E$ in order to enforce a monotonic structure on the estimated out-of-sample variances.
\end{remark}


\subsection{Related Literature}
    Our proposed estimator corresponds to a single choice of a split location (that is, the size of the training fold), which we denote as  $M$. The approach of \cite{abadir2014design} considers averaging the estimated eigenvalues over different fold $k$  as well as the split location $M \in \mathcal{M}$ where $\mathcal{M}=[0.2T, 0.8T]$ is the range of split locations. Their estimator is then given by the following grand average
    \begin{align}
        \hat{\Sigma} &= \sum _{M\in \mathcal{M}} \hat{\Sigma}_M , \label{eq:split} 
    \end{align}
    where $\hat{\Sigma}_M \equiv \hat{\Sigma}$ is the estimator \eqref{eq:reconstr} with $W=\mathbb{I}_T$ for a given split location $M$. This estimator is asymptotically optimal if the split location satisfies $M$, $T-M \to \infty$ with $M/T \to c \in (0, 1)$; see \citet[Proposition 3]{abadir2014design}. 
    
    In contrast, \cite{lam2016nonparametric} considers a similar setup but studies the large-dimensional setting where $N \to \infty$ as $T \to \infty$ such that $N/T \to c > 0$. Instead of averaging over a range of splits, \citet[Theorem 5]{lam2016nonparametric} proposed a split $M = T - aT^{1/2}$, with some constant $a >0$, that is asymptotically efficient. However, this asymptotic property may be challenged in finite samples or as $M/T$ approaches a constant smaller than 1, and there is no practical guide for setting the constant $a$. \citet[Section 4.2]{lam2016nonparametric} proposes to search $M$ over seven candidate split locations $\mathcal{M} = [2T^{1/2}, 0.2T, 0.4T, 0.6T, 0.8T,T - 2.5T^{1/2},T - 1.5T^{1/2}]$ in order to minimize a criterion similar to that of \cite{bickel2008regularized}.  
    
    In our paper, we avoid the non-trivial task of choosing the split locations entirely following \cite{bartz} to reduce the number of required computations to build the estimator.

\subsection{Comparison to DCC-NL}
Another estimator that is often compared to a covariance matrix measured through an exponentially weighted scheme is the Dynamic Conditional Correlation (DCC) model proposed by \cite{engle2002dynamic} to model the time-varying dynamics of multivariate financial returns due to conditional heteroskedasticity. As is well known, the DCC model also suffers from the `curse of dimensionality'. There have been efforts from \cite{engle2019large} to successfully enable the estimation of the DCC to be both (i) computationally feasible in the large dimensions through the use of composite likelihood functions from \cite{pakel2021fitting}, and (ii) with reduced estimation error through the nonlinear shrinkage (NL) of \cite{ledoit2015spectrum}. This estimator is referred to as the DCC-NL. 

Both EWA-CV and DCC-NL are intimately related by the concept of shrinkage but they also differ in a few other ways.\footnote{\cite{lam2016nonparametric} proved the data splitting scheme of \cite{abadir2014design} is a nonparametric approach of achieving a similar nonlinear shrinkage effect in \cite{ledoit2012nonlinear}. } First, the latter explicitly applies devolatization to the returns. Therefore, in order to allow for a more fair comparison between both estimators, EWA-CV should also be applied to a panel of returns series that have been devolatized through a univariate GARCH model of \cite{bollerslev1986generalized}. Second, the dynamic nature of DCC requires that the covariance be determined recursively, while the EWA-SC requires only matrix multiplications. Third, the DCC-NL is governed by two parameters, while EWA-CV has just one. Our view is that maintaining a single parameter can offer a more parsimonious representation of asset returns, which provides insights  into the sensitivity of the estimator's performance to only changes in the decay rate.  Moreover, since the decay rate is typically specified by the user in practice, it possesses less implementation overhead over the DCC, whose dynamic parameters are typically estimated via maximum likelihood. Admittedly, EWA-CV is a numerical scheme compared to NL, an analytical method, but it only employs a few spectral decompositions and vectorized operations to estimate an $N$-dimensional vector, which are fast to execute. Finally, EWA-CV is conceptually simpler since it relies on the principle of cross-validation, which many practitioners are familiar with, whereas the DCC-NL requires recourse to the theory of random matrices. Notwithstanding the similarities and differences between both approaches, we formally compare their performances in Section \ref{sec:empiricalchp1}. 


\section{Simulation Study}\label{sec:montecarlo}




\subsection{Return Generating Process}\label{sec:dgp}
We now study the effectiveness of our proposed estimator against different competing estimators through a challenging simulation environment, where the risk dynamics are driven by the RiskMetrics model of \cite{rm96}.\footnote{ The RiskMetrics 1996 methodology is a covariance matrix that evolves recursively according to an exponentially weighted moving average. It has a similar form to our EWA-SC \eqref{eq:emaSC} up to timing in the measurement.} This allows us to assess if our proposed estimator is accurate enough to be useful in an ideal environment where it is supposed to thrive. In particular, we consider a return-generating process (RGP) that is drawn from a multivariate standard normal distribution given by
\begin{align}
    x_t = \Sigma_t ^{1/2} z_t \quad \text{with} \quad z_t \stackrel{\text{i.i.d.}}{\sim} \mathcal{N} (\mathbf{0}, \mathbb{I}_N),
\end{align}
where $\Sigma_t$ is a population covariance matrix that evolves according to
\begin{align}\label{eq:rm94}
    \Sigma_t := \beta _\Sigma \Sigma_{t-1} + (1-\beta _\Sigma) x_{t-1} x_{t-1} ' . 
\end{align}
The parameter $\beta _\Sigma$ is the `intrinsic' decay rate of the population covariance matrix, which we set to $0.996$ in this simulation study. We use this RGP to generate a data set of $N=500$ asset returns with $T=1250$ daily return observations, which corresponds to roughly 5 years of daily data.

\subsection{Candidate Estimators}

From the generated data set of asset returns, we can estimate a covariance matrix. Given the plethora of covariance estimators in the literature, we shall limit our analysis to the following candidate estimators for $\hat{\Sigma}$, which are either uniformly weighted or exponentially weighted:
\begin{itemize}[label={\tiny\raisebox{1ex}{\textbullet}}]
    \item LS: the linear shrinkage estimator of \cite{ledoit2004well}, with the shrinkage target being (a scaled multiple of) the identity matrix.
    \item NL: the quadratic inverse shrinkage estimator of \cite{ledoit2022quadratic}.
    \item CV: the cross-validation shrinkage estimator of \cite{bartz} with $K=10$ folds.
    \item EWA-SC: the estimator from Equation \eqref{eq:emaSC}.
    \item EWA-CV: based on our proposed cross-validation shrinkage estimator from Equation \eqref{eq:reconstr} with $K=10$ folds. 
\end{itemize}
For EWA-based estimators, we shall consider various specifications of decay rates $\beta$ ranging from $0.999$ to $0.99$. This would allow us to see how they behave under model misspecification.

\subsection{Performance Assessment}

To compare the performance of various covariance estimators against the population covariance, we need a loss function and an evaluation metric. The loss between an estimator $\hat{\Sigma}$ and the population covariance matrix $\Sigma$ is based on Equation \eqref{eq:loss}. 

A simple metric to quantify an improvement of an estimator over a benchmark model is the so-called, PRIAL (percentage of relative improvement in average loss). If we let the standard sample covariance be the benchmark model, then the PRIAL is defined as
\begin{align} \label{eq:prial}
    \text{PRIAL}(\hat{\Sigma}, S) := \left ( 1 - \frac{ \mathbb{E} [\mathcal{L}(\hat{\Sigma}, \Sigma _T)]}{\mathbb{E}[\mathcal{L}(S, \Sigma _T)]} \right ) \times 100 \% ,
\end{align}
where $\Sigma _T$ is a population covariance obtained at the end of the sample period, and $\mathcal{L}$ is a loss function. The PRIAL is $100 \%$ when our estimator coincides with the population covariance $\Sigma_T$ at time $T$, and $0\%$ with the uniformly-weighted sample covariance $S$. A negative value for PRIAL is possible, which indicates that the estimator did not improve over the reference matrix $S$. The notation $\mathbb{E}[\cdot]$ denotes an expected value of the losses, which is taken to be the average over the 100 simulation trials.

\subsection{Simulation Results}
\begin{figure}[H]
\caption{Out-of-Sample Risk Improvement of Different Estimators}
    \captionsetup{font=footnotesize}
\caption*{The figures display the PRIAL of several candidate covariance estimators with respect to the standard sample covariance. The left panel compares uniformly-weighted and exponentially-weighted estimators over different decay rates $\beta$. The return generating process is stated in Section \ref{sec:dgp}. The vertical dashed line at $0.996$ corresponds to the intrinsic decay rate of population covariance. The average is taken over $100$ repetitions. The right panel compares EWA-CV for the different number of folds $K$ under the same setup.}
    \begin{subfigure}[b]{0.49\textwidth}
         \centering
         \includegraphics[width=\textwidth]{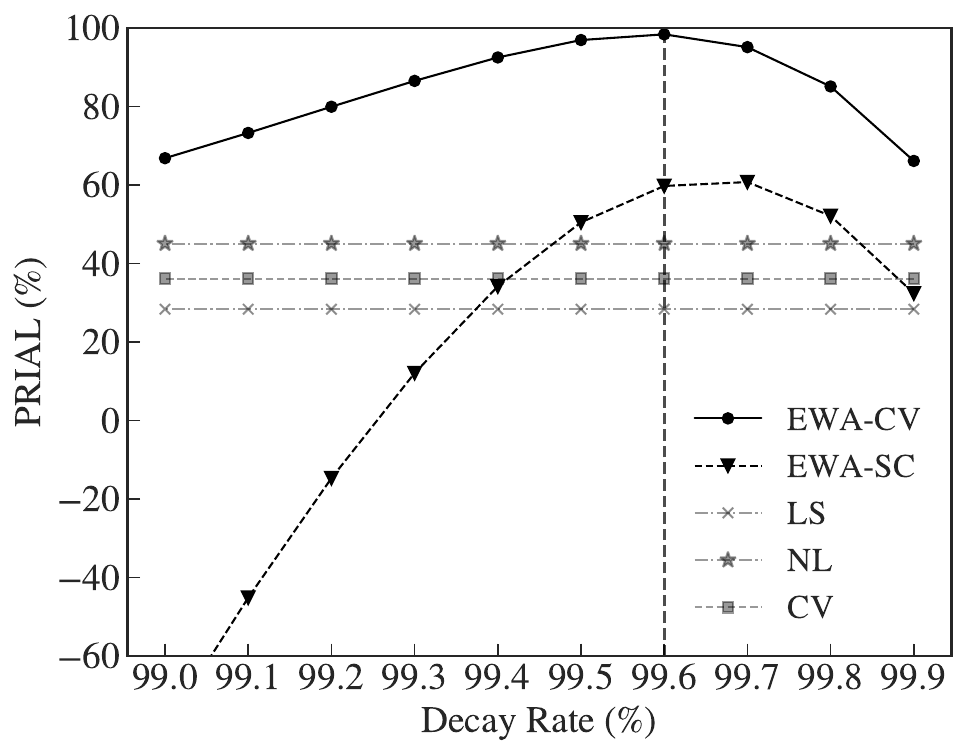}
     \end{subfigure}
     \begin{subfigure}[b]{0.49\textwidth}
         \centering
         \includegraphics[width=\textwidth]{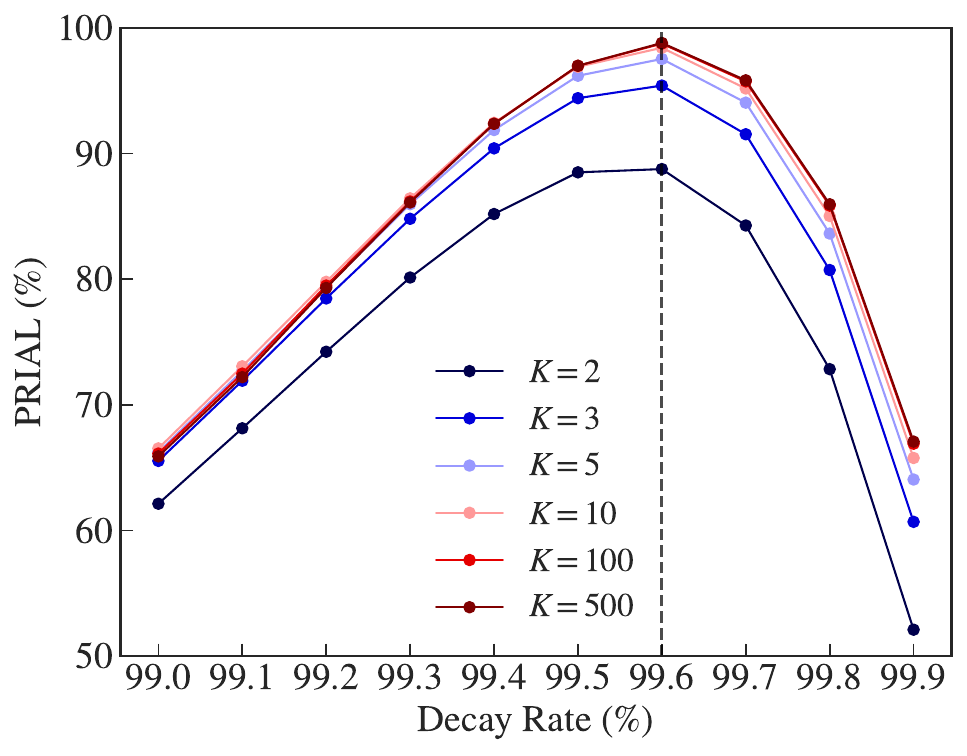}
     \end{subfigure}
    \label{fig:risk}
\end{figure}

We now present the results of the Monte Carlo simulations. Figure \ref{fig:risk} shows the PRIAL for different covariance estimators against the different decay rate specifications. Starting with the uniformly weighted-based estimators, we see that LS, NL, and CV perform better than the standard sample covariance $S$ as seen by their positive PRIAL values. This improvement owes to the benefits of shrinkage. However, their performances remain flat for all decay rate specifications since, by their uniformly weighted construction, they are not parameterized by them. This lack of flexibility of the uniformly-weighted estimators fundamentally limits their ability to align themselves toward more recent risk, which gives rise to higher estimation errors.

In contrast, EWA-based estimators beat uniformly weighted estimators when their decay rate $\beta$ is in the vicinity around the intrinsic decay rate at $\beta _\Sigma =0.996$ with EWA-CV performing the best having PRIAL values of greater than 90\%. This is reassuring for us because the EWA-based estimators \eqref{eq:emaSC} have the same form as the RGP in \eqref{eq:rm94} (up to timing in measurement). Moreover, this also speaks to the importance of capturing time series variation of risk since the EWA-based estimators differ from their uniformly weighted counterparts by how the observations are measured across time. 

A further inspection into the performance of EWA-SC shows that there is very little margin for error in specifying $\beta$. In particular, EWA-SC starts to rapidly underperform all other estimators as $\beta$ decreases and even takes on negative values for $\beta$ less than $0.993$. Fortunately, our proposed EWA-CV does not share that fragile behavior. The shrinkage technology that we have introduced in Section \ref{sec:emacv} has allowed it to be robust against a large ensemble of noisy observations causing it to yield stellar outperformance even with a misspecified decay rate. This is a desirable feature because, in practice, the actual decay rate of an ensemble of non-stationary return series is not known and we want to be protected from placing bets on risky sample eigenvectors by mistake. 


Finally, we ran this experiment on a single server with an (256GB RAM 48 cores) Intel Xeon Gold 6252N 2.3GHz CPU. To estimate a 500-dimensional covariance of this setup, EWA-CV takes about 0.5 seconds wall-clock time and, by way of comparison, NL takes about 0.1 seconds. Thus, our method is fast to execute and incurs negligible runtime overhead.

\subsection{Sensitivity Analysis to the Number of Folds}

We also conduct a sensitivity analysis of EWA-CV to the different number of folds $K$. Figure \ref{fig:risk} shows that overall, the performance of EWA-CV is not highly sensitive to the number of folds. While its performance can be sub-optimal for $K < 10$, it is consistently optimal for $K \geq 10$. The intuition is that for $K < 10$, there is not enough averaging across the folds to further suppress the noise from the projections of the test observation onto the sample eigenvectors. However, for $K \geq 10$, we find that the performances remain similar. In light of these findings, we will use $K=10$ folds for all of our experiments, which is sufficient for us to reap the most benefits from averaging and to strike an appropriate balance between computational tractability and precision.

\section{Empirical Analysis}\label{sec:empiricalchp1}

\subsection{Data and Portfolio Construction Rules} \label{sec:data}

We now analyze the out-of-sample performance of our proposed estimator using real financial data. We first obtain daily stock returns data from the Center for Research in Security Prices (CRSP), starting at $01/01/1981$ and ending at $12/31/2019$. Our analysis is limited to stocks traded on the NYSE, AMEX, and NASDAQ stock exchanges. The size of the investment universe that we consider is $N \in \{ 100, 500, 1000 \}$.  

We define one `month' and `year' as $21$ and $252$ consecutive trading days, respectively. The portfolios are rebalanced on a monthly basis using a rolling window scheme where only past information is used to avoid a look-ahead bias. During the month, the number of shares is kept fixed in order to avoid incurring unnecessary transaction costs. The covariance matrices are estimated using an in-sample period of size $T = 1250$, which is approximately 5 years. The out-of-sample period starts from $09/04/1986$ and ends at $12/31/2019$. This provides us with a total of $m=400$ months (or $8400$ trading days) of consecutive nonoverlapping observations of length $21$ days for which the portfolios are rebalanced. 

To construct a well-defined investment universe on which we can estimate the covariances on, we use a similar procedure described in \cite{engle2019large}. For each rebalancing month $h = 0, 1, \ldots, m-1$ (using zero-based indexing), we first select the stocks that have complete data over the $T=1250$ days in the in-sample period as well as over the $21$ days in the out-of-sample period. The backward and forward restrictions ensure that we have data to estimate our models and to evaluate out-of-sample. Then, we search for pairs of highly correlated stocks (that is, those with a sample correlation exceeding $0.95$) and remove the stock with the lower volume in each pair. From this remaining set of stocks, we select the largest $N$ stocks as measured by their market capitalization on the rebalancing month $h$ to include in our investment universe. 

\subsection{Base Candidate Portfolios}
To test the performance of different covariance matrix estimators in the context of portfolio selection, we consider two investment problems: (1) the global minimum variance (GMV) portfolio and (2) the Markowitz portfolio with a given signal. Both problems are described in Section \ref{sec:gmv} and Appendix \ref{sec:mps}, respectively. In either setting, we employ the following candidate estimators as their inputs:
\begin{itemize}[label={\tiny\raisebox{1ex}{\textbullet}}]
    \item SC: the sample covariance matrix \eqref{eq:SC}.
     \item LS: the linear shrinkage estimator of \cite{ledoit2004honey}, with the shrinkage target being (a scaled multiple of) the identity matrix.
    \item NL: the quadratic inverse shrinkage (QIS) estimator of \cite{ledoit2022quadratic}.
    \item EWA-SC: the exponentially-weighted average sample covariance \eqref{eq:emaSC}.
    \item EWA-CV: our proposed method based on estimator \eqref{eq:reconstr} with $K=10$. 
    \item AFM1-NL: an approximate single factor model of \cite{de2021factor} with NL applied to the residuals.
    \item AFM1-EWA-CV: our second proposed method based on an approximate single factor model with EWA-CV applied to the residuals. 
\end{itemize}
Note that for the EWA-based estimators, we choose a fixed decay rate that is relatively high at $\beta = 0.997$, which makes them slightly closer to the uniformly weighted setting. While this choice might seem \textit{ad-hoc}, nonetheless, our robustness checks show that it is a suitable choice for our financial application. 

\begin{remark}[Factor models]
     While many active portfolio managers tend to hold relatively concentrated portfolios of less than 100 stocks, there are quantitative portfolio managers that invest in a broad investment universe spanning thousands of single stock equities in order to benefit from efficiency gains in large dimensions. In such a case, they tend to exploit some form of factor structure in order to make the investment problem more manageable. To accommodate for this practice, we can also extend our proposed estimator towards data that have been generated by a factor model in a similar spirit to \cite{de2021factor} but with EWA–CV applied to the residual terms. In particular, we will utilize the market factor, which is easily accessible to researchers, and term our second proposed covariance estimator in this paper AFM1-EWA-CV. 
\end{remark}

\subsection{Evaluation Methodology}
To evaluate the performance of the different portfolios, we report four main out-of-sample performance measures: the average out-of-sample returns (AV), the standard deviation of portfolio returns (SD), the information ratio (IR), and the Sharpe ratio (SR). For ease of interpretability, all performance measures are annualized with 252 trading ‘days’. The information ratio is computed with respect to the actual returns (as opposed to returns in excess of the risk-free rate) since we believe it is more relevant in our context where the portfolios are formed solely on the basis of risky assets. On the other hand, the Sharpe ratio is the information ratio with the risk-free rate as the benchmark. In addition, we also include the maximum drawdown (MDD), which measures the largest cumulative loss from peak to trough over the entire out-of-sample period. 

We also report the information ratio and Sharpe ratio net transaction costs denoted as $\widetilde{\mathrm{IR}}$ and $\widetilde{\mathrm{SR}}$, respectively. In particular, we assume a constant transaction cost of 5 bps for all stocks, which is a reasonable approximation today. Even though the investment problem that we consider below does not explicitly account for transaction costs at the portfolio pre-formation stage, it is nevertheless worthwhile to investigate if our methods survive transaction costs in an unconstrained setup.


Additionally, we report the following portfolio weight statistics averaged over the 400 trading months: turnover (TO), gross exposure (GE; computed as the sum of the absolute value of long and short positions), proportional leverage (PL; computed as the fraction of negative weights), minimum weight (MIN), maximum weight (MAX), the standard deviation of weights (SD), and mean absolute deviation from an equally-weighted portfolio (MAD). Note that these statistics are not our primary focus since our proposed method is not optimized to account for these measures. Nevertheless, they are provided to give a better overview of the different methods.

\subsection{Global Minimum Variance Portfolio}\label{sec:gmv}
The global minimum variance (GMV) portfolio is expressed as
\begin{equation}\label{eq:gmv}
\begin{aligned}
    &\min_{w} w' \Sigma w\\
    \text{such that} \quad & \begin{aligned} 
            &w' \mathbf{1} = 1,
            \end{aligned}
\end{aligned}
\end{equation}
where $w = (w_1 , \ldots , w_N)'$ is a vector of portfolio weights. The GMV portfolio allocation framework is useful for isolating the estimation error stemming from the expected returns and allows us to focus solely on evaluating the quality of a covariance matrix estimator. This application is similar to \cite{engle2019large} for reproducibility. Empirical studies of \cite{haugen1991efficient} and \cite{jagannathan2003risk}, among others, have found that the GMV portfolios enjoy favorable out-of-sample performance both in terms of risk and Sharpe ratio when compared with other benchmark portfolios. Moreover, it is also adopted in practice, as is the case, for example, with the iShares Edge MSCI Min Vol USA ETF, which has seen high asset inflows since early 2013; see for example, \cite{goldberg2013restoring}. 

In this setting, the solution for optimal weights has an explicit form given by 
\begin{align}\label{eq:minport}
    w = \frac{\Sigma^{-1} \mathbf{1}}{ \mathbf{1}'\Sigma^{-1} \mathbf{1}} ,
\end{align}
where $\mathbf{1}$ is an $N$-dimensional conformable vector of ones. To yield a feasible portfolio denoted as $\hat{w}$ that can be computed from the returns data, we replace $\Sigma$ with the covariance matrix estimator $\hat{\Sigma}$. We also include the equal-weighted portfolio promoted by \cite{dgu} as another benchmark, which is denoted by $1/N$, since it is argued to be difficult to beat.

The primary metric for which we evaluate the performance of the GMV portfolio will be the out-of-sample standard deviation. We also report the statistical significance of the differences between the SD values of EWA-SC and EWA-CV derived by using the two-sided p-value of the prewhitened $\mathrm{HAC}_{\mathrm{PW}}$ test described by \citet[Section 3.1]{ledoit2011robust}. Since the purpose of this paper is to demonstrate that EWA-CV is an improvement over EWA-SC, we confine our attention to the comparison of EWA-SC with EWA-CV to avoid multiple testing problems. 

The out-of-sample performances are reported in Table \ref{tab:gmvperf} and the results from the SD column are summarized as follows: 
\begin{itemize}[label={\tiny\raisebox{1ex}{\textbullet}}]
    \item  EWA-CV outperforms EWA-SC for all $N$ and is statistically significant at the 0.01 level for $N=500, 1000$.
    \item Overall, the EWA-CV portfolio delivers the best results, owing in part to its ability to adapt to the non-stationary nature of the financial returns. Its performance can be seen to improve with increasing universe size $N$.
    \item The EWA-SC portfolio is competitive for $N=100$ and even outperforms portfolios that are regularized with shrinkage. This is probably due to the fact that at relatively small dimensions, a covariance estimator may benefit more from picking up more recent time series variations. However, as $N$ increases, the EWA-SC portfolio starts to underperform. This also corroborates the empirical findings of \citet[Table 7]{engle2019large} in their application of the RiskMetrics 2006 methodology of \cite{zumbach}, which is a more elaborate version to the EWA-SC but also inherits the same cross-sectional estimation errors. 
    \item Even though AFM1-EWA-CV has slightly better performance than the NL estimators, it does not lead to any material improvement above EWA-CV.
    \item The reduction in the standard deviation of EWA-CV over NL in $N=100, 500, 1000$ amounts to 5\%, 4\%, and 2\%, respectively. While the outperformance for $N=1000$ may not be statistically significant at the 0.01 level, it is still an improvement and our method offers a viable alternative to the analytical nonlinear shrinkage method. A similar conclusion holds for EWA-CV and AFM1-NL since both AFM1-NL and NL share similar performance, which is consistent with the findings of \citet[Table A.1]{de2021factor}. 
\end{itemize}

Additionally, we find that EWA-CV outperforms in terms of information ratio and Sharpe ratio for $N=1000$ although the performance between the portfolios is similar for $N=100, 500$. Once we have taken into account the transaction costs at the post-formation stage of the portfolio, we see a decline in the information and Sharpe ratios across all portfolios but the relative performance remain similar. In terms of the maximum drawdown incurred by the portfolio, we see that both EWA-CV and AFM1-EWA-CV suffer the least. For instance, if $N = 1000$, the maximum drawdown for $1/N$ is 55.04\%, while for EWA-CV, it is 25.63\%, which is a 50\% reduction.

Table \ref{tab:gmvweight} describes the distribution of the portfolio weights of the estimated portfolios. We see that EWA-CV has the least dispersed weight while both SC and EWA-SC have the most dispersed weight.  Furthermore, EWA-CV and AFM1-EWA-CV have the lowest turnover, which is interesting given that we make no effort to control the trajectory of the weights. Moreover, EWA-CV also has the lowest amount of gross exposure and proportion of leverage.

\subsection{Robustness Checks}

In this section, we inspect whether the outperformance of our proposed estimator is robust to different revisions in the current empirical set-up. In particular, we will be interested in parameters dictating the (1) subsample period, (2) exponential decay rate, (3) no short-sale constraint, and (4) devolatization.

\subsubsection{Subsample Analysis}

Here, we check if there are any peculiar subsample effects that may drive the performances of our proposed scheme. We divide the out-of-sample period into four equally-sized subsamples of 100 months (that is, roughly 8 trading years) each: (1) 09/04/1986 to 12/21/1994, (2) 12/22/1994 to 04/25/2003, (3) 04/28/2003 to 08/25/2011, and (4) 08/26/2011 to 12/31/2019. Then we perform the same procedure in each subsample. The results are provided in Table \ref{tab:gmvSubsample}. 

We see that the ranking of performance across different portfolios is relatively consistent over time. Generally, EWA-CV is the best performer although there are brief episodes where it is not. For example, for $N=100$, EWA-SC outperforms in Sample 1, while for $N=1000$, NL (and AFM1-NL) outperforms in Sample 4. Nevertheless, the performance of EWA-CV is close to the leading ones in these scenarios.

\subsubsection{Sensitivity Analysis}

In our base experiments, we parameterized the EWA-based estimators with a decay rate of $0.997$. We now repeat the same backtest exercise as in the previous section but focus only on EWA-based estimators for different exponential decay rate specifications $\beta \in \{ 0.999, 0.997, 0.995, 0.992, 0.990, 0.985 \}$ and $N\in \{ 100, 1000 \}$. This allows us to examine if the performance of our estimator is sensitive to changes in the decay rate specification or if there are other choices that may lead to better results.

Table \ref{tab:gmvsensitive} shows that the outperformance of EWA-CV over EWA-SC is robust across different decay rates for $N=1000$ while for $N=100$, it only marginally outperforms EWA-CV in Sample 1 for $\beta \geq 0.99$. Moreover, EWA-CV tends to deliver a lower standard deviation with decay rates greater than $0.995$. Unreported simulation results also indicate that decay rates lower than 0.985 consistently give a higher standard deviation. This observation holds over the full sample period as well as across different subsamples. These empirical findings are certainly at odds with the recommended value of 0.94 suggested by \cite{rm96} for modeling exponentially weighted moving average covariances.

\subsubsection{No Short-Sale Constraint}
In our original investment problem \eqref{eq:gmv}, we allow for the possibility of short-selling. However, shorting in the equity markets is difficult and a 50\% margin must be posted. Since some portfolio managers may encounter a no-short-sale constraint in practice, we now impose a non-negative constraint on the portfolio weights in the optimization objective.

Table \ref{tab:gmvnss} presents the results. As can be seen, SC and EWA-SC are now competitive in terms of standard deviation, even though it differs only slightly from other estimators. By comparing these results to Table \ref{tab:gmvperf}, it is evident that prohibiting short sales benefits SC and EWA-SC but harms other shrinkage estimators, including EWA-CV. Based on \cite{jagannathan2003risk}, these results concur with them that enforcing a no-short-sales constraint has an implicit shrinkage effect on the sample covariances when estimating the global minimum variance portfolio.

\subsubsection{Devolatization}

The base case experiments focus exclusively on the performance improvement that exponential weighting adds over a uniformly weighted shrinkage estimator. We now consider applying EWA-based estimators to a panel of returns series that have been devolatized through a univariate GARCH model. For this effort, we use Kevin Sheppard's Python $\code{arch}$ package to implement the GARCH(1,1) model.

We are now in a position to compare against the DCC-NL model of \cite{engle2019large} as well as the AFM1-DCC-NL model of \cite{de2021factor}. To this end, we use the Matlab implementation of DCC-NL found on Michael Wolf's website but replaced the correlation targeting matrix with the QIS estimator.\footnote{The code can be downloaded at \url{http://www.econ.uzh.ch/en/people/faculty/wolf/publications.html}.} Note that if at any rebalancing date, the estimation of the GARCH model (via the Matlab routine) produces a convergence issue, we add to the returns a noise term that is normally distributed with an expected value of 0 and a level of standard deviation, which increases from $10^{-8}$ to $10^{-4}$ in step size of 20 according to a geometric progression until convergence is achieved.

Finally, given that we exclude one of the two assets whose correlations are in excess of 0.95, our investment universe includes only stocks with `moderate' correlations. In such an environment, it may be the case that having an accurate correlation matrix estimate is of second-order importance relative to estimating the variances of each asset accurately. Thus, as another benchmark, we consider a GMV portfolio, which assumes a diagonal correlation matrix with the variances estimated with a GARCH(1,1) model, denoted as VOL.

Table \ref{tab:gmvdevol} summarizes the performances of the five portfolios: VOL, DCC-NL, EWA-CV, AFM1-DCC-NL, and AFM1-EWA-CV.  Compared to Table \ref{tab:gmvperf}, the standard deviation of EWA-CV estimators has reduced with devolatization for $N=500, 1000$. For example, for $N=1000$, the standard deviation for EWA-CV and AFM1-EWA-CV decreased by 9.7\% and 16.4\%, respectively. Moreover, AFM1-EWA-CV appears to benefit the most from devolatization and is now the best performer for $N = 500, 1000$. 

We also see that the performance of EWA-CV estimators is better than VOL for all $N$ by a wide margin. Hence, the performance improvements we observe in our estimator are not just due to better estimates of individual stocks' volatility; they also stem from better estimates of correlations.

Interestingly, the performances of EWA-CV and DCC-NL estimators are quite similar for all investment sizes $N$. For instance, for $N=1000$, the standard deviation reduction of EWA-CV over DCC-NL is 3\%, whereas for AFM1-EWA-CV over AFM1-DCC-NL it is 1\%. This observation is not unexpected given that both estimators employ a similar dynamic capture and shrinkage technology.  The EWA-CV estimators also tend to be accompanied by a marginal improvement in average returns compared to their DCC-NL analogs, and also slightly better gross and net information and Sharpe ratios.

\section{Conclusion}\label{sec:conclusionchp1}

This paper introduces a novel covariance estimator that performs in large dimensions and adapts to the non-stationarities or persistent heteroskedastic environments of financial stock returns. Such scenarios are often found in practical applications of portfolio optimization for which our estimator is highly relevant. We accomplish this goal by (1) weighting the sample covariance matrix towards more recent data with exponentially weighted averages (EWA), and (2) through a novel application of the cross-validation (CV) shrinkage technology of \cite{abadir2014design}, \cite{lam2016nonparametric} and \cite{bartz}. The former allows us to capture the non-stationary dynamics of financial time series with more flexible sample eigenvectors, while the latter attenuates the adverse noise effects that are embedded in the sample eigenvalues. We call our estimator, EWA-CV. Beyond these statistical benefits, our proposed estimator possesses several appealing features such as the simplicity of implementation, low runtime overhead for research prototyping, and transparency for independent verification. 

We have provided simulation experiments to help us to gain insights into the channels that drive the outperformance of our estimator over the sample  exponentially weighted average covariance and other uniformly weighted estimators. Our empirical experiments on real financial data reinforce its superiority over the exponentially weighted sample covariance and uniformly weighted shrinkage estimators with improved realized measures in large investment universe sizes. We also demonstrated that our proposed shrinkage estimator(s) rivals the performance of the state-of-the-art dynamic conditional correlation non-linear shrinkage (DCC-NL) estimator of \cite{engle2019large} and its factor model extension by \cite{de2021factor} while incurring less implementation and computational overhead as well as simpler routines. Hence, our proposal offers practitioners of risk and asset management an attractive reduced-form alternative to the existing large-dimensional multivariate dynamic volatility models for financial returns. 


In general, the adverse noise effects from a large-dimensional regime can be further reduced by incorporating additional prior knowledge about the orientation of the eigenvectors. For example, this can include overlaying the covariance matrix with economically driven models that possess a small number of risk parameters or with an appropriately structured block matrix reflecting the sector relationships of the stock returns. A cross-pollination of a postulated covariance matrix with the technology of nonlinear shrinkage provides an interesting avenue for further research. 

We hope that EWA-CV will serve as a valuable addition to a portfolio manager's armory and a new benchmark in large cross-sectional financial applications.

\newpage

\bibliographystyle{apalike}
\bibliography{papers}  



\appendix

\newpage
\clearpage
\pagestyle{empty}

\newpage
\section{Figures}\label{appendix:figureschp1}

\newpage
\section{Tables}\label{appendix:tableschp1}

\vspace*{\fill}
\begin{table}[H]
\caption{Performance Summary}
    \captionsetup{font=footnotesize}
    \caption*{This table shows the annualized out-of-sample performances (in percentages) of global minimum variance portfolios of various covariance estimators. The columns display the average returns (AV), the standard deviation of returns (SD), the information ratio (IR), the Sharpe ratio (SR), and the maximum drawdown (MDD). The information ratio and Sharpe ratio net of transaction costs of 5 basis points are also reported and denoted as $\widetilde{\mathrm{IR}}$ and $\widetilde{\mathrm{SR}}$, respectively. Panels A-C display the results for investment universe sizes $N \in \{ 100, 500, 1000 \}$.  The daily out-of-sample returns are from $09/04/1986$ up to $12/31/2019$. The most competitive value in the SD column is highlighted in \textbf{bold}. For the EWA-SC and EWA-CV portfolios, an asterisk indicates a significant outperformance of one portfolio versus the other in terms of SD: $^{***}$ denotes significance at the 0.01 level; $^{**}$ denotes significance at the 0.05 level; and $^{*}$ denotes significance at the 0.1 level.}
\centering
\small
\begin{tabular}{lccccccc} \toprule
 & AV & SD & IR & SR & $\widetilde{\mathrm{IR}}$ & $\widetilde{\mathrm{SR}}$ & MDD\\ \cmidrule{1-8}
\multicolumn{8}{l}{Panel A: Universe Size, $N = 100$} \\\cmidrule{1-8}
$1/N$ & $12.20$ & $18.08$ & $0.67$ & $0.51$ & $0.67$ & $0.51$ & $54.48$\\ 
SC & $9.60$ & $11.86$ & $0.81$ & $0.55$ & $0.79$ & $0.53$ & $38.11$\\ 
EWA-SC & $8.45$ & $11.37$ & $0.74$ & $0.47$ & $0.71$ & $0.44$ & $37.66$\\ 
LS & $9.77$ & $11.79$ & $0.83$ & $0.57$ & $0.81$ & $0.55$ & $36.48$\\ 
NL & $9.67$ & $11.75$ & $0.82$ & $0.56$ & $0.80$ & $0.54$ & $36.96$\\ 
EWA-CV & $9.00$ & $\textbf{11.17}^{*}$ & $0.81$ & $0.53$ & $0.78$ & $0.51$ & $36.03$\\ 
AFM1-NL & $9.66$ & $11.76$ & $0.82$ & $0.56$ & $0.80$ & $0.54$ & $36.90$\\ 
AFM1-EWA-CV & $9.28$ & $11.57$ & $0.80$ & $0.54$ & $0.78$ & $0.51$ & $34.84$\\ 
\cmidrule{1-8}
\multicolumn{8}{l}{Panel B: Universe Size, $N =500$} \\\cmidrule{1-8}
$1/N$ & $12.63$ & $17.68$ & $0.71$ & $0.54$ & $0.71$ & $0.54$ & $55.47$\\ 
SC & $9.13$ & $8.80$ & $1.04$ & $0.69$ & $0.94$ & $0.59$ & $34.24$\\ 
EWA-SC & $8.48$ & $9.08$ & $0.93$ & $0.60$ & $0.80$ & $0.46$ & $39.77$\\ 
LS & $9.43$ & $8.39$ & $1.12$ & $0.76$ & $1.05$ & $0.68$ & $32.38$\\ 
NL & $8.99$ & $7.94$ & $1.13$ & $0.75$ & $1.08$ & $0.70$ & $30.15$\\ 
EWA-CV & $8.77$ & $\textbf{7.58}^{***}$ & $1.16$ & $0.75$ & $1.11$ & $0.70$ & $25.54$\\ 
AFM1-NL & $8.93$ & $7.94$ & $1.13$ & $0.74$ & $1.08$ & $0.69$ & $30.54$\\ 
AFM1-EWA-CV & $8.86$ & $7.78$ & $1.14$ & $0.75$ & $1.09$ & $0.70$ & $27.47$\\ 
\cmidrule{1-8}
\multicolumn{8}{l}{Panel C: Universe Size, $N = 1000$} \\\cmidrule{1-8}
$1/N$ & $12.83$ & $17.40$ & $0.74$ & $0.56$ & $0.74$ & $0.56$ & $55.04$\\ 
SC & $6.92$ & $9.93$ & $0.70$ & $0.39$ & $0.37$ & $0.07$ & $34.47$\\ 
EWA-SC & $6.91$ & $10.46$ & $0.66$ & $0.37$ & $0.32$ & $0.03$ & $35.83$\\ 
LS & $8.10$ & $7.11$ & $1.14$ & $0.71$ & $0.99$ & $0.56$ & $33.56$\\ 
NL & $9.01$ & $6.51$ & $1.38$ & $0.92$ & $1.31$ & $0.84$ & $27.81$\\ 
EWA-CV & $9.46$ & $\textbf{6.39}^{***}$ & $1.48$ & $1.00$ & $1.42$ & $0.94$ & $25.63$\\ 
AFM1-NL & $8.96$ & $6.51$ & $1.38$ & $0.91$ & $1.30$ & $0.83$ & $27.82$\\ 
AFM1-EWA-CV & $9.56$ & $6.51$ & $1.47$ & $1.00$ & $1.41$ & $0.94$ & $27.11$\\ 
\bottomrule \end{tabular}
    \label{tab:gmvperf}
\end{table}
\vspace{\fill}

\newpage

\vspace*{\fill}
\begin{table}[H]
\caption{Weight Statistics}
    \captionsetup{font=footnotesize}
    \caption*{This table shows the average monthly weight statistics of global minimum variance portfolios of various covariance estimators. The columns display the turnover (TO), the gross exposure (GE), the proportional leverage (PL), the minimum weight (MIN), the maximum weight (MAX), the standard deviation of weights (SD), and the mean absolute deviation from an equally-weighted portfolio (MAD). Panels A-C display the results for investment universe sizes $N \in \{ 100, 500, 1000 \}$. The summary statistics are obtained from an average over the 400 portfolio formation periods. }
\centering
\small
\begin{tabular}{lccccccc} \toprule
 & TO & GE & PL & MIN & MAX & SD & MAD\\ \cmidrule{1-8}
\multicolumn{8}{l}{Panel A: Universe Size, $N = 100$} \\\cmidrule{1-8}
$1/N$ & $0.064$ & $1.000$ & $0.000$ & $0.010$ & $0.010$ & $0.000$ & $0.000$\\ 
SC & $0.626$ & $2.956$ & $0.442$ & $-0.083$ & $0.253$ & $0.045$ & $0.030$\\ 
EWA-SC & $0.882$ & $3.164$ & $0.442$ & $-0.094$ & $0.249$ & $0.047$ & $0.032$\\ 
LS & $0.572$ & $2.772$ & $0.429$ & $-0.075$ & $0.193$ & $0.040$ & $0.028$\\ 
NL & $0.548$ & $2.713$ & $0.428$ & $-0.072$ & $0.217$ & $0.040$ & $0.027$\\ 
EWA-CV & $0.663$ & $2.613$ & $0.410$ & $-0.069$ & $0.180$ & $0.037$ & $0.026$\\ 
AFM1-NL & $0.552$ & $2.727$ & $0.429$ & $-0.073$ & $0.221$ & $0.041$ & $0.027$\\ 
AFM1-EWA-CV & $0.643$ & $2.715$ & $0.416$ & $-0.074$ & $0.206$ & $0.039$ & $0.027$\\ 
\cmidrule{1-8}
\multicolumn{8}{l}{Panel B: Universe Size, $N =500$} \\\cmidrule{1-8}
$1/N$ & $0.052$ & $1.000$ & $0.000$ & $0.002$ & $0.002$ & $0.000$ & $0.000$\\ 
SC & $1.670$ & $5.191$ & $0.475$ & $-0.070$ & $0.255$ & $0.021$ & $0.010$\\ 
EWA-SC & $2.322$ & $5.935$ & $0.476$ & $-0.081$ & $0.276$ & $0.023$ & $0.012$\\ 
LS & $1.284$ & $4.475$ & $0.459$ & $-0.036$ & $0.138$ & $0.015$ & $0.009$\\ 
NL & $0.802$ & $3.399$ & $0.437$ & $-0.046$ & $0.201$ & $0.015$ & $0.007$\\ 
EWA-CV & $0.812$ & $3.141$ & $0.424$ & $-0.044$ & $0.191$ & $0.014$ & $0.006$\\ 
AFM1-NL & $0.811$ & $3.425$ & $0.439$ & $-0.045$ & $0.203$ & $0.015$ & $0.007$\\ 
AFM1-EWA-CV & $0.815$ & $3.223$ & $0.427$ & $-0.043$ & $0.196$ & $0.014$ & $0.006$\\ 
\cmidrule{1-8}
\multicolumn{8}{l}{Panel C: Universe Size, $N = 1000$} \\\cmidrule{1-8}
$1/N$ & $0.048$ & $1.000$ & $0.000$ & $0.001$ & $0.001$ & $0.000$ & $0.000$\\ 
SC & $5.693$ & $9.734$ & $0.486$ & $-0.087$ & $0.235$ & $0.017$ & $0.010$\\ 
EWA-SC & $6.368$ & $10.453$ & $0.486$ & $-0.089$ & $0.240$ & $0.018$ & $0.010$\\ 
LS & $2.029$ & $5.521$ & $0.465$ & $-0.023$ & $0.077$ & $0.008$ & $0.006$\\ 
NL & $1.025$ & $3.517$ & $0.426$ & $-0.037$ & $0.128$ & $0.007$ & $0.003$\\ 
EWA-CV & $0.845$ & $3.113$ & $0.407$ & $-0.024$ & $0.089$ & $0.006$ & $0.003$\\ 
AFM1-NL & $1.037$ & $3.536$ & $0.427$ & $-0.036$ & $0.129$ & $0.007$ & $0.004$\\ 
AFM1-EWA-CV & $0.857$ & $3.185$ & $0.410$ & $-0.024$ & $0.091$ & $0.006$ & $0.003$\\ 
\bottomrule \end{tabular}
    \label{tab:gmvweight}
\end{table}
\vspace{\fill}

\newpage

\vspace*{\fill}
\begin{table}[H]
\caption{Subsample Analysis}
    \captionsetup{font=footnotesize}
    \caption*{This table shows the annualized out-of-sample  standard deviation and information ratio of global minimum variance portfolios of various covariance estimators over different subsamples. The columns display the subsample periods divided equally into Sample 1 (1986 through 1994), Sample 2 (1994 through 2003), Sample 3 (2003 through 2011), and Sample 4 (2011 through 2019). Panels A-C display the results for investment universe sizes $N \in \{ 100, 500, 1000 \}$. The daily out-of-sample returns are from $09/04/1986$ up to $12/31/2019$.  The most competitive value in the SD column is highlighted in \textbf{bold}.}
\centering
\small
\begin{tabular}{lccccccccccc} \toprule
& \multicolumn{2}{c}{Sample 1} &  & \multicolumn{2}{c}{Sample 2} &  & \multicolumn{2}{c}{Sample 3} &  & \multicolumn{2}{c}{Sample 4} \\ \cline{2-3} \cline{5-6} \cline{8-9} \cline{11-12}& SD & IR && SD & IR && SD & IR && SD & IR\\ \cmidrule{1-12}
\multicolumn{12}{l}{Panel A: Universe Size, $N = 100$} \\\cmidrule{1-12}
$1/N$ & $16.62$ & $0.71$&  & $19.72$ & $0.77$&  & $21.25$ & $0.36$&  & $13.84$ & $1.02$\\ 
SC & $11.74$ & $0.74$&  & $14.00$ & $0.54$&  & $11.54$ & $0.97$&  & $9.81$ & $1.12$\\ 
EWA-SC & $\textbf{10.91}$ & $0.83$&  & $13.38$ & $0.43$&  & $11.41$ & $0.82$&  & $9.46$ & $1.02$\\ 
LS & $11.67$ & $0.75$&  & $13.94$ & $0.54$&  & $11.44$ & $1.00$&  & $9.74$ & $1.16$\\ 
NL & $11.66$ & $0.76$&  & $13.86$ & $0.52$&  & $11.44$ & $0.99$&  & $9.68$ & $1.17$\\ 
EWA-CV & $11.09$ & $0.81$&  & $\textbf{13.07}$ & $0.45$&  & $\textbf{11.00}$ & $0.91$&  & $\textbf{9.19}$ & $1.20$\\ 
AFM1-NL & $11.68$ & $0.76$&  & $13.86$ & $0.52$&  & $11.45$ & $0.98$&  & $9.68$ & $1.16$\\ 
AFM1-EWA-CV & $11.40$ & $0.79$&  & $13.77$ & $0.42$&  & $11.17$ & $1.01$&  & $9.57$ & $1.16$\\ 
\cmidrule{1-12}
\multicolumn{12}{l}{Panel B: Universe Size, $N =500$} \\\cmidrule{1-12}
$1/N$ & $14.80$ & $0.81$&  & $17.86$ & $0.76$&  & $22.75$ & $0.50$&  & $13.96$ & $0.97$\\ 
SC & $9.02$ & $1.11$&  & $9.59$ & $1.08$&  & $10.32$ & $0.84$&  & $5.49$ & $1.38$\\ 
EWA-SC & $8.90$ & $1.03$&  & $10.28$ & $1.07$&  & $10.76$ & $0.66$&  & $5.43$ & $1.22$\\ 
LS & $8.52$ & $1.14$&  & $9.16$ & $1.11$&  & $9.77$ & $0.89$&  & $5.49$ & $1.66$\\ 
NL & $8.01$ & $1.14$&  & $8.92$ & $0.97$&  & $9.20$ & $1.08$&  & $4.89$ & $1.68$\\ 
EWA-CV & $\textbf{7.41}$ & $1.17$&  & $\textbf{8.73}$ & $0.97$&  & $\textbf{8.83}$ & $1.13$&  & $\textbf{4.54}$ & $1.75$\\ 
AFM1-NL & $8.00$ & $1.13$&  & $8.92$ & $0.96$&  & $9.19$ & $1.08$&  & $4.90$ & $1.66$\\ 
AFM1-EWA-CV & $7.68$ & $1.16$&  & $9.01$ & $0.91$&  & $8.96$ & $1.17$&  & $4.64$ & $1.70$\\ 
\cmidrule{1-12}
\multicolumn{12}{l}{Panel C: Universe Size, $N = 1000$} \\\cmidrule{1-12}
$1/N$ & $13.46$ & $0.89$&  & $16.90$ & $0.77$&  & $23.13$ & $0.55$&  & $14.49$ & $0.94$\\ 
SC & $9.75$ & $0.67$&  & $9.71$ & $0.92$&  & $12.21$ & $0.56$&  & $7.48$ & $0.72$\\ 
EWA-SC & $10.37$ & $0.62$&  & $10.45$ & $0.75$&  & $12.76$ & $0.67$&  & $7.64$ & $0.63$\\ 
LS & $6.70$ & $1.08$&  & $6.80$ & $1.47$&  & $8.83$ & $0.75$&  & $5.75$ & $1.48$\\ 
NL & $6.08$ & $1.32$&  & $6.73$ & $1.43$&  & $8.12$ & $1.15$&  & $\textbf{4.61}$ & $1.96$\\ 
EWA-CV & $\textbf{5.84}$ & $1.32$&  & $\textbf{6.69}$ & $1.45$&  & $\textbf{7.90}$ & $1.22$&  & $4.72$ & $2.29$\\ 
AFM1-NL & $6.08$ & $1.32$&  & $6.72$ & $1.42$&  & $8.12$ & $1.16$&  & $4.61$ & $1.93$\\ 
AFM1-EWA-CV & $5.97$ & $1.33$&  & $6.81$ & $1.37$&  & $8.02$ & $1.25$&  & $4.79$ & $2.28$\\ 
\bottomrule \end{tabular}
    \label{tab:gmvSubsample}
\end{table}
\vspace{\fill}

\newpage

\vspace*{\fill}
\begin{table}[H]
\caption{Sensitivity Analysis}
    \captionsetup{font=footnotesize}
    \caption*{This table shows the annualized out-of-sample standard deviation (in percentages) and information ratio of global minimum variance portfolios estimated with EWA-SC and EWA-CV for different decay rates across different subsamples. The columns display the full sample period and the subsample periods divided equally into Sample 1 (1986 through 1994), Sample 2 (1994 through 2003), Sample 3 (2003 through 2011), and Sample 4 (2011 through 2019). Panels A-B and C-D display the results for investment universe sizes $N=100$ and $N=1000$, respectively.  The daily out-of-sample returns are from $09/04/1986$ up to $12/31/2019$. }
\centering
\small
\begin{tabular}{lcccccccccccccc} \toprule
& \multicolumn{2}{c}{Full Sample} &  &\multicolumn{2}{c}{Sample 1} &  & \multicolumn{2}{c}{Sample 2} &  & \multicolumn{2}{c}{Sample 3} &  & \multicolumn{2}{c}{Sample 4} \\ \cline{2-3} \cline{5-6} \cline{8-9} \cline{11-12} \cline{14-15}& SD & IR && SD & IR && SD & IR && SD & IR && SD & IR\\ \cmidrule{1-15}
\multicolumn{15}{l}{Panel A: EWA-SC, Universe Size: $N=100$} \\\cmidrule{1-15}
$\beta =0.999$ & $11.56$ & $0.80$&  & $11.30$ & $0.78$&  & $13.62$ & $0.51$&  & $11.36$ & $0.93$&  & $9.61$ & $1.08$\\ 
$\beta =0.997$ & $11.37$ & $0.74$&  & $10.91$ & $0.83$&  & $13.38$ & $0.43$&  & $11.41$ & $0.82$&  & $9.46$ & $1.02$\\ 
$\beta =0.995$ & $11.55$ & $0.68$&  & $10.94$ & $0.83$&  & $13.53$ & $0.34$&  & $11.77$ & $0.70$&  & $9.61$ & $1.01$\\ 
$\beta =0.992$ & $12.03$ & $0.60$&  & $11.24$ & $0.76$&  & $14.03$ & $0.22$&  & $12.41$ & $0.59$&  & $10.07$ & $0.99$\\ 
$\beta =0.990$ & $12.38$ & $0.55$&  & $11.48$ & $0.70$&  & $14.44$ & $0.15$&  & $12.81$ & $0.55$&  & $10.41$ & $0.96$\\ 
$\beta =0.985$ & $13.27$ & $0.44$&  & $12.11$ & $0.57$&  & $15.57$ & $-0.00$&  & $13.73$ & $0.47$&  & $11.26$ & $0.90$\\ 
\cmidrule{1-15}
\multicolumn{15}{l}{Panel B: EWA-CV, Universe Size: $N=100$} \\\cmidrule{1-15}
$\beta =0.999$ & $11.43$ & $0.82$&  & $11.30$ & $0.81$&  & $13.46$ & $0.48$&  & $11.17$ & $0.98$&  & $9.44$ & $1.16$\\ 
$\beta =0.997$ & $11.17$ & $0.81$&  & $11.09$ & $0.81$&  & $13.07$ & $0.45$&  & $11.00$ & $0.91$&  & $9.19$ & $1.20$\\ 
$\beta =0.995$ & $11.15$ & $0.80$&  & $11.20$ & $0.78$&  & $12.94$ & $0.44$&  & $11.00$ & $0.86$&  & $9.13$ & $1.28$\\ 
$\beta =0.992$ & $11.23$ & $0.80$&  & $11.43$ & $0.75$&  & $12.90$ & $0.40$&  & $11.11$ & $0.83$&  & $9.15$ & $1.40$\\ 
$\beta =0.990$ & $11.28$ & $0.80$&  & $11.53$ & $0.72$&  & $12.95$ & $0.40$&  & $11.15$ & $0.83$&  & $9.16$ & $1.45$\\ 
$\beta =0.985$ & $11.40$ & $0.78$&  & $11.72$ & $0.64$&  & $13.03$ & $0.42$&  & $11.28$ & $0.81$&  & $9.24$ & $1.49$\\ 
\cmidrule{1-15}
\multicolumn{15}{l}{Panel C: EWA-SC, Universe Size: $N=1000$} \\\cmidrule{1-15}
$\beta =0.999$ & $9.97$ & $0.69$&  & $9.83$ & $0.67$&  & $9.80$ & $0.87$&  & $12.25$ & $0.58$&  & $7.41$ & $0.71$\\ 
$\beta =0.997$ & $10.46$ & $0.66$&  & $10.37$ & $0.62$&  & $10.45$ & $0.75$&  & $12.76$ & $0.67$&  & $7.64$ & $0.63$\\ 
$\beta =0.995$ & $11.35$ & $0.63$&  & $11.23$ & $0.51$&  & $11.47$ & $0.65$&  & $13.76$ & $0.75$&  & $8.28$ & $0.58$\\ 
$\beta =0.992$ & $12.94$ & $0.57$&  & $12.73$ & $0.34$&  & $13.13$ & $0.57$&  & $15.63$ & $0.81$&  & $9.54$ & $0.54$\\ 
$\beta =0.990$ & $14.00$ & $0.54$&  & $13.75$ & $0.26$&  & $14.22$ & $0.53$&  & $16.86$ & $0.82$&  & $10.41$ & $0.53$\\ 
$\beta =0.985$ & $16.46$ & $0.47$&  & $16.15$ & $0.13$&  & $16.83$ & $0.42$&  & $19.66$ & $0.76$&  & $12.39$ & $0.53$\\ 
\cmidrule{1-15}
\multicolumn{15}{l}{Panel D: EWA-CV, Universe Size: $N=1000$} \\\cmidrule{1-15}
$\beta =0.999$ & $6.45$ & $1.48$&  & $5.92$ & $1.35$&  & $6.68$ & $1.45$&  & $7.99$ & $1.22$&  & $4.77$ & $2.23$\\ 
$\beta =0.997$ & $6.39$ & $1.48$&  & $5.84$ & $1.32$&  & $6.69$ & $1.45$&  & $7.90$ & $1.22$&  & $4.72$ & $2.29$\\ 
$\beta =0.995$ & $6.48$ & $1.46$&  & $5.91$ & $1.25$&  & $6.83$ & $1.40$&  & $7.95$ & $1.24$&  & $4.83$ & $2.26$\\ 
$\beta =0.992$ & $6.69$ & $1.40$&  & $6.15$ & $1.13$&  & $7.04$ & $1.34$&  & $8.10$ & $1.25$&  & $5.11$ & $2.14$\\ 
$\beta =0.990$ & $6.82$ & $1.37$&  & $6.32$ & $1.08$&  & $7.16$ & $1.32$&  & $8.19$ & $1.24$&  & $5.28$ & $2.09$\\ 
$\beta =0.985$ & $7.13$ & $1.31$&  & $6.71$ & $0.97$&  & $7.45$ & $1.28$&  & $8.45$ & $1.18$&  & $5.62$ & $2.05$\\ 
\bottomrule \end{tabular}
    \label{tab:gmvsensitive}
\end{table}
\vspace{\fill}

\newpage

\vspace*{\fill}
\begin{table}[H]
\caption{No Short-Sale Constraint}
    \captionsetup{font=footnotesize}
    \caption*{This table shows the annualized out-of-sample performances (in percentages) of global minimum variance portfolios of various covariance estimators.  Short sales are prohibited due to a lower bound of zero on all portfolio weights. The columns display the average returns (AV), the standard deviation of returns (SD), the information ratio (IR), the Sharpe ratio (SR), and the maximum drawdown (MDD). The information ratio and Sharpe ratio net of transaction costs of 5 basis points are also reported and denoted as $\widetilde{\mathrm{IR}}$ and $\widetilde{\mathrm{SR}}$, respectively. Panels A-C display the results for investment universe sizes $N \in \{ 100, 500, 1000 \}$. The daily out-of-sample returns are from $09/04/1986$ up to $12/31/2019$. The most competitive value in the SD column is highlighted in \textbf{bold}. For the EWA-SC and EWA-CV portfolios, an asterisk indicates a significant outperformance of one portfolio versus the other in terms of SD: $^{***}$ denotes significance at the 0.01 level; $^{**}$ denotes significance at the 0.05 level; and $^{*}$ denotes significance at the 0.1 level.}
\centering
\small
\begin{tabular}{lccccccc} \toprule
 & AV & SD & IR & SR & $\widetilde{\mathrm{IR}}$ & $\widetilde{\mathrm{SR}}$ & MDD\\ \cmidrule{1-8}
\multicolumn{8}{l}{Panel A: Universe Size, $N = 100$} \\\cmidrule{1-8}
$1/N$ & $12.20$ & $18.08$ & $0.67$ & $0.51$ & $0.67$ & $0.51$ & $54.48$\\ 
SC & $9.83$ & $12.61$ & $0.78$ & $0.54$ & $0.77$ & $0.53$ & $40.81$\\ 
EWA-SC & $9.46$ & $12.04$ & $0.79$ & $0.53$ & $0.78$ & $0.52$ & $38.25$\\ 
LS & $9.86$ & $12.60$ & $0.78$ & $0.54$ & $0.78$ & $0.53$ & $40.35$\\ 
NL & $9.80$ & $12.60$ & $0.78$ & $0.54$ & $0.77$ & $0.53$ & $40.84$\\ 
EWA-CV & $9.51$ & $\textbf{12.10}$ & $0.79$ & $0.53$ & $0.78$ & $0.53$ & $38.65$\\ 
AFM1-NL & $9.80$ & $12.61$ & $0.78$ & $0.54$ & $0.77$ & $0.53$ & $40.92$\\ 
AFM1-EWA-CV & $9.78$ & $12.57$ & $0.78$ & $0.54$ & $0.77$ & $0.53$ & $38.60$\\ 
\cmidrule{1-8}
\multicolumn{8}{l}{Panel B: Universe Size, $N =500$} \\\cmidrule{1-8}
$1/N$ & $12.63$ & $17.68$ & $0.71$ & $0.54$ & $0.71$ & $0.54$ & $55.47$\\ 
SC & $10.39$ & $9.86$ & $1.05$ & $0.74$ & $1.05$ & $0.74$ & $39.10$\\ 
EWA-SC & $10.22$ & $\textbf{9.23}$ & $1.11$ & $0.78$ & $1.09$ & $0.76$ & $38.97$\\ 
LS & $10.46$ & $9.87$ & $1.06$ & $0.75$ & $1.05$ & $0.74$ & $38.72$\\ 
NL & $10.13$ & $9.91$ & $1.02$ & $0.71$ & $1.02$ & $0.71$ & $38.98$\\ 
EWA-CV & $9.77$ & $9.33$ & $1.05$ & $0.72$ & $1.04$ & $0.71$ & $37.92$\\ 
AFM1-NL & $10.12$ & $9.92$ & $1.02$ & $0.71$ & $1.01$ & $0.71$ & $39.06$\\ 
AFM1-EWA-CV & $9.80$ & $9.83$ & $1.00$ & $0.69$ & $0.99$ & $0.68$ & $38.00$\\ 
\cmidrule{1-8}
\multicolumn{8}{l}{Panel C: Universe Size, $N = 1000$} \\\cmidrule{1-8}
$1/N$ & $12.83$ & $17.40$ & $0.74$ & $0.56$ & $0.74$ & $0.56$ & $55.04$\\ 
SC & $8.72$ & $7.76$ & $1.12$ & $0.73$ & $1.11$ & $0.72$ & $31.69$\\ 
EWA-SC & $8.22$ & $\textbf{7.32}$ & $1.12$ & $0.71$ & $1.11$ & $0.69$ & $34.00$\\ 
LS & $9.12$ & $7.85$ & $1.16$ & $0.77$ & $1.15$ & $0.76$ & $32.66$\\ 
NL & $9.51$ & $8.27$ & $1.15$ & $0.78$ & $1.14$ & $0.77$ & $34.34$\\ 
EWA-CV & $9.26$ & $7.82$ & $1.18$ & $0.79$ & $1.17$ & $0.78$ & $33.64$\\ 
AFM1-NL & $9.49$ & $8.27$ & $1.15$ & $0.78$ & $1.14$ & $0.77$ & $34.52$\\ 
AFM1-EWA-CV & $9.26$ & $8.36$ & $1.11$ & $0.74$ & $1.10$ & $0.73$ & $35.96$\\ 
\bottomrule \end{tabular}
    \label{tab:gmvnss}
\end{table}
\vspace{\fill}

\newpage

\vspace*{\fill}
\begin{table}[H]
\caption{Devolatization}
    \captionsetup{font=footnotesize}
    \caption*{This table shows the annualized out-of-sample performances (in percentages) of global minimum variance portfolios for various estimators. The columns display the average returns (AV), the standard deviation of returns (SD), the information ratio (IR), the Sharpe ratio (SR), and the maximum drawdown (MDD). The information ratio and Sharpe ratio net of transaction costs of 5 basis points are also reported and denoted as $\widetilde{\mathrm{IR}}$ and $\widetilde{\mathrm{SR}}$, respectively. Panels A-C display the results for investment universe sizes $N \in \{ 100, 500, 1000 \}$. The daily out-of-sample returns are from $09/04/1986$ up to $12/31/2019$. The most competitive value in the SD column is highlighted in \textbf{bold}.}
\centering
\small
\begin{tabular}{lccccccc} \toprule
 & AV & SD & IR & SR & $\widetilde{\mathrm{IR}}$ & $\widetilde{\mathrm{SR}}$ & MDD\\ \cmidrule{1-8}
\multicolumn{8}{l}{Panel A: Universe Size, $N = 100$} \\\cmidrule{1-8}
VOL & $11.88$ & $15.41$ & $0.77$ & $0.57$ & $0.76$ & $0.56$ & $45.63$\\
DCC-NL & $9.18$ & $11.69$ & $0.78$ & $0.52$ & $0.67$ & $0.41$ & $38.91$\\ 
EWA-CV & $10.63$ & $11.57$ & $0.92$ & $0.65$ & $0.82$ & $0.56$ & $40.00$\\ 
AFM1-DCC-NL & $9.02$ & $\textbf{11.52}$ & $0.78$ & $0.52$ & $0.72$ & $0.45$ & $37.03$\\ 
AFM1-EWA-CV & $9.35$ & $11.60$ & $0.81$ & $0.54$ & $0.74$ & $0.47$ & $32.39$\\ 
\cmidrule{1-8}
\multicolumn{8}{l}{Panel B: Universe Size, $N =500$} \\\cmidrule{1-8}
VOL & $11.47$ & $13.68$ & $0.84$ & $0.62$ & $0.82$ & $0.60$ & $47.56$\\ 
DCC-NL & $9.10$ & $7.48$ & $1.22$ & $0.81$ & $1.05$ & $0.65$ & $27.89$\\ 
EWA-CV & $9.25$ & $7.19$ & $1.29$ & $0.86$ & $1.15$ & $0.73$ & $30.13$\\ 
AFM1-DCC-NL & $9.22$ & $7.27$ & $1.27$ & $0.85$ & $1.14$ & $0.72$ & $29.83$\\ 
AFM1-EWA-CV & $9.49$ & $\textbf{7.13}$ & $1.33$ & $0.90$ & $1.21$ & $0.78$ & $28.32$\\ 
\cmidrule{1-8}
\multicolumn{8}{l}{Panel C: Universe Size, $N = 1000$} \\\cmidrule{1-8}
VOL & $11.33$ & $12.91$ & $0.88$ & $0.64$ & $0.86$ & $0.63$ & $48.20$\\ 
DCC-NL & $8.20$ & $5.95$ & $1.38$ & $0.87$ & $1.20$ & $0.69$ & $29.78$\\ 
EWA-CV & $8.30$ & $5.77$ & $1.44$ & $0.91$ & $1.29$ & $0.77$ & $33.10$\\ 
AFM1-DCC-NL & $8.22$ & $5.50$ & $1.49$ & $0.94$ & $1.34$ & $0.79$ & $24.23$\\ 
AFM1-EWA-CV & $8.45$ & $\textbf{5.44}$ & $1.55$ & $0.99$ & $1.41$ & $0.85$ & $27.02$\\ 
\bottomrule \end{tabular}
    \label{tab:gmvdevol}
\end{table}
\vspace{\fill}

\newpage

\section{Markowitz Portfolio with Signal}\label{sec:mps}
The Markowitz problem with a fully-invested constraint is expressed as
\begin{align}\label{eq:mvoobj}
    \begin{aligned} 
        &\min _{w} w' \Sigma w \\
         \text{such that} \quad & \begin{aligned} 
            &w' m \geq b, \enskip \text{and} \\
            &w' \mathbf{1} = 1,
            \end{aligned}
\end{aligned} 
\end{align}
where $m$ is a predictive signal and $b$ is an investor's minimum target expected return. We consider the momentum signal as there is substantial empirical evidence that documents this anomaly in the returns of individual stocks \citep{jegadeesh1990evidence, jegadeesh1993returns}. A momentum signal is calculated by taking the geometric average of the past year's stock returns, excluding the most recent month's returns. Taking the current momentum values for all $N$ stocks in the universe produces a cross-sectional predictive signal $m$. This application is similar to \cite{engle2019large} for reproducibility.

The well-known solution to this optimization problem is given by
\begin{equation} \label{eq:mvo:chp1}
    \begin{aligned}
        w &= c_1 \Sigma^{-1} \mathbf{1} + c_2 \Sigma^{-1} m, \\
        \text{where } c_1 &:= \frac{C - b B}{AC - B^2}, \quad \text{and} \quad c_2 := \frac{bA - B}{AC - B^2}, \\
        \text{with } A &:= \mathbf{1}' \Sigma^{-1} \mathbf{1}, \quad B := \mathbf{1}'\Sigma^{-1} m, \quad \text{and} \quad C := m'\Sigma^{-1} m  ,
    \end{aligned}
\end{equation}
where $\mathbf{1}$ is a vector of ones. A feasible portfolio $\hat{w}$ can be obtained by replacing the unknown covariance $\Sigma$ with an estimator $\hat{\Sigma}$ in Equation \eqref{eq:mvo:chp1}. We also include an equally-weighted portfolio of the top quintile stocks sorted according to their momentum value, which is labeled as EW-TQ. Further, we set the target expected return $b$ to be the average of the momentum values of the stocks from in this portfolio.


%

The primary metric for which we evaluate the performance of Markowitz portfolios will be the out-of-sample information ratio. We also report the statistical significance of the differences between the IR values of EWA-SC and EWA-CV derived by using the two-sided p-value of the prewhitened $\mathrm{HAC}_{\mathrm{PW}}$ test described by \citet[Section 3.1]{ledoit2008robust}. Since the purpose of this paper is to demonstrate that EWA-CV is an improvement over EWA-SC, we confine our attention to the comparison of EWA-SC with EWA-CV to avoid a multiple testing problem.

The out-of-sample performances are reported in Table \ref{tab:mvoperf} and the results from the IR column are summarized as follows: 

\begin{itemize}[label={\tiny\raisebox{1ex}{\textbullet}}]
    \item EWA-CV outperforms EWA-SC for all $N$ and is statistically and economically significant at the 0.01 level for $N=500, 1000$.
    \item Overall, the EWA-CV portfolio delivers the best results, owing in part to its ability to adapt to the non-stationary nature of the financial returns. Its performance can be seen to improve with increasing universe size $N$.
    \item Even though AFM1-EWA-CV has slightly better performance than the NL estimators, it does not lead to any material improvement above EWA-CV.
    \item The increase in information ratio of EWA-CV over NL in $N=100, 500, 1000$ amounts to 1\%, 6\%, and 9\%, respectively. While this outperformance may not be statistically significant at the 0.01 level or economically meaningful, it is still an improvement and our method offers a viable alternative to the analytical nonlinear shrinkage method. A similar conclusion holds for the comparison between EWA-CV and AFM1-NL since both AFM1-NL and NL share similar performance, which is consistent with the findings of \citet[Table A.3]{de2021factor}. 
\end{itemize}

In addition, we find that EWA-CV produces the highest out-of-sample information ratio and Sharpe ratio after incorporating transaction costs at the post-formation stage of the portfolio for $N = 500, 1000$. Further, \cite{engle2006testing} suggests assessing the performance of a Markowitz portfolio in terms of the out-of-sample standard deviation. Based on this metric, we find that EWA-CV has the lowest value across different $N$. In terms of the maximum drawdown incurred by the portfolio, we see that both EWA-CV and AFM1-EWA-CV suffer the least for $N = 500, 1000$. For instance, if $N = 1000$, the maximum drawdown for EW-TQ is 58.56\%, while for EWA-CV, it is 23.19\%, which is a 60\% reduction.

Table \ref{tab:mvoweight} describes the distribution of the portfolio weights of the estimated portfolios. We see that EWA-CV has the least dispersed weight while both SC and EWA-SC have the most dispersed weight.  Furthermore, EWA-CV and AFM1-EWA-CV have the lowest turnover, which is interesting given that we make no effort to control the trajectory of the weights. Moreover, EWA-CV also has the lowest amount of gross exposure and proportion of leverage.

\subsection{Robustness Checks}

In this section, we inspect whether the outperformance of our proposed estimator is robust to different revisions in the current empirical set-up. In particular, we will be interested in parameters dictating the (1) subsample period, (2) exponential decay rate, (3) no short-sale constraint, and (4) devolatization.

\subsubsection{Subsample Analysis}

Here, we check if there are any peculiar subsample effects that may drive the performances of our proposed scheme. We divide the out-of-sample period into four equally-sized subsamples of 100 months (that is, roughly 8 trading years) each: (1) 09/04/1986 to 12/21/1994, (2) 12/22/1994 to 04/25/2003, (3) 04/28/2003 to 08/25/2011, and (4) 08/26/2011 to 12/31/2019. Then we perform the same procedure in each subsample. The results are provided in Table \ref{tab:mvoSubsample}. 

We see that the ranking of performance across different portfolios is relatively consistent over time for $N=500, 1000$ with EWA-CV estimators being the best performers. For $N=100$, however, there are brief episodes where it is not; EWA-SC outperforms in Sample 1, EW-TQ in Sample 2, and NL (and AFM1-NL) in Sample 4. Nevertheless, the performance of EWA-CV is close to the leading ones in these scenarios. In terms of standard deviation, EWA-CV consistently has the lowest value across all subsamples and universes.

\subsubsection{Sensitivity Analysis}

In our base experiments, we parameterized the EWA-based estimators with a decay rate of $0.997$. We now repeat the same backtest exercise as in the previous section but focus only on EWA-based estimators for different exponential decay rate specifications $\beta \in \{ 0.999, 0.997, 0.995, 0.992, 0.990, 0.985 \}$ and $N\in \{ 100, 1000 \}$. This allows us to examine if the performance of our estimator is sensitive to changes in the decay rate specification or if there are other choices that may lead to better results.

According to Table \ref{tab:mvosensitive}, EWA-CV outperforms EWA-SC across different decay rates for $N = 1000$, whereas for $N = 100$, EWA-SC outperforms EWA-CV only marginally in Sample 1. Moreover, for $N=1000$, EWA-CV tends to deliver a higher information ratio with decay rates greater than $0.995$. Unreported simulation results also indicate that decay rates lower than 0.985 consistently give a lower information ratio.  This observation holds over the full sample period as well as across different subsamples. For $N=100$, there appears to be some performance benefit from tilting the EWA-CV towards more recent observations in some periods. For example, in Sample 2 and Sample 4, $\beta = 0.985$ gives the highest information ratio, which is a 23\% and 15\% improvement over using $\beta = 0.997$.



\subsubsection{No Short-Sale Constraint}
In our original investment problem \eqref{eq:mvoobj}, we allow for the possibility of short-selling. Since some portfolio managers may encounter a no-short-sale constraint in practice, we now impose a non-negative constraint on the portfolio weights in the optimization objective.

Table \ref{tab:mvonss} presents the results. As can be seen, SC and EWA-SC are now competitive in terms of information ratio, even though it differs only slightly from other estimators. By comparing these results to Table \ref{tab:mvoperf}, it is evident that prohibiting short sales benefits SC and EWA-SC but harms other shrinkage estimators, including EWA-CV. 

\subsubsection{Devolatization}

The base case experiments focus exclusively on the performance improvement exponential weighting adds over a uniform weighted shrinkage estimator. We now consider applying EWA-based estimators to a panel of returns series that have been devolatized through a univariate GARCH model. This will allow us to facilitate comparison with the DCC-NL model of \cite{engle2019large}. As another benchmark, we consider a Markowitz portfolio, which assumes a diagonal correlation matrix with the variances estimated with a GARCH(1,1) model, denoted as VOL. 



Table \ref{tab:mvodevol}  summarizes the performances of the five portfolios: VOL, DCC-NL, EWA-CV, AFM1- DCC-NL, and AFM1-EWA-CV. Compared to Table \ref{tab:mvoperf}, it can be seen that the information ratio of EWA-CV has increased with devolatization for $N = 100, 500$ but reduced for $N=1000$. AFM1-EWA-CV appears to benefit the most from devolatization and is now the best performer for $N = 500, 1000$.

We also see that the information ratios of EWA-CV estimators are higher than VOL for all $N$ by a wide margin. Hence, the performance improvements we observe in our estimator are not just due to better estimates of individual stocks' volatility; they also stem from better estimates of correlations.

Interestingly, the performances of EWA-CV and DCC-NL estimators are quite similar for all investment sizes $N$. For instance, the information ratio improvement of EWA-CV over DCC-NL for $N=1000$ is 4\%, whereas for AFM1-EWA-CV over AFM1-DCC-NL it is 1\%. EWA-CV estimators also have slightly better net information and Sharpe ratios. 

\newpage

\vspace*{\fill}
\begin{table}[H]
\caption{Performance Summary}
    \captionsetup{font=footnotesize}
    \caption*{This table shows the annualized out-of-sample performances (in percentages) of Markowitz portfolios of various covariance estimators. The columns display the average returns (AV), the standard deviation of returns (SD), the information ratio (IR), the Sharpe ratio (SR), and the maximum drawdown (MDD). The information ratio and Sharpe ratio net of transaction costs of 5 basis points are also reported and denoted as $\widetilde{\mathrm{IR}}$ and $\widetilde{\mathrm{SR}}$, respectively. Panels A-C display the results for investment universe sizes $N \in \{ 100, 500, 1000 \}$.  The daily out-of-sample returns are from $09/04/1986$ up to $12/31/2019$. The most competitive value in the IR column is highlighted in \textbf{bold}. For the EWA-SC and EWA-CV portfolios, an asterisk indicates a significant outperformance of one portfolio versus the other in terms of IR: $^{***}$ denotes significance at the 0.01 level; $^{**}$ denotes significance at the 0.05 level; and $^{*}$ denotes significance at the 0.1 level.}
    \centering
    \small
\centering
\small
\begin{tabular}{lccccccc} \toprule
 & AV & SD & IR & SR & $\widetilde{\mathrm{IR}}$ & $\widetilde{\mathrm{SR}}$ & MDD\\ \cmidrule{1-8}
\multicolumn{8}{l}{Panel A: Universe Size, $N = 100$} \\\cmidrule{1-8}
EW-TQ & $15.62$ & $22.30$ & $0.70$ & $0.56$ & $0.69$ & $0.55$ & $56.71$\\ 
SC & $12.19$ & $14.47$ & $0.84$ & $0.63$ & $0.80$ & $0.59$ & $36.83$\\ 
EWA-SC & $10.98$ & $14.01$ & $0.78$ & $0.57$ & $0.73$ & $0.51$ & $41.63$\\ 
LS & $12.40$ & $14.41$ & $0.86$ & $0.65$ & $0.82$ & $0.61$ & $36.60$\\ 
NL & $12.46$ & $14.42$ & $0.86$ & $0.65$ & $0.83$ & $0.61$ & $37.25$\\ 
EWA-CV & $11.98$ & $13.85$ & $\textbf{0.87}^{**}$ & $0.64$ & $0.82$ & $0.60$ & $41.32$\\ 
AFM1-NL & $12.47$ & $14.42$ & $0.86$ & $0.65$ & $0.83$ & $0.61$ & $36.98$\\ 
AFM1-EWA-CV & $12.11$ & $14.19$ & $0.85$ & $0.64$ & $0.81$ & $0.60$ & $40.41$\\ 
\cmidrule{1-8}
\multicolumn{8}{l}{Panel B: Universe Size, $N =500$} \\\cmidrule{1-8}
EW-TQ & $14.69$ & $21.44$ & $0.69$ & $0.54$ & $0.67$ & $0.53$ & $58.63$\\ 
SC & $11.28$ & $10.36$ & $1.09$ & $0.79$ & $0.97$ & $0.67$ & $33.83$\\ 
EWA-SC & $9.96$ & $10.88$ & $0.92$ & $0.64$ & $0.76$ & $0.48$ & $36.06$\\ 
LS & $11.84$ & $9.99$ & $1.18$ & $0.88$ & $1.09$ & $0.78$ & $32.23$\\ 
NL & $12.00$ & $9.67$ & $1.24$ & $0.93$ & $1.17$ & $0.86$ & $29.03$\\ 
EWA-CV & $12.16$ & $9.24$ & $\textbf{1.32}^{***}$ & $0.99$ & $1.24$ & $0.91$ & $23.39$\\ 
AFM1-NL & $11.94$ & $9.67$ & $1.24$ & $0.92$ & $1.17$ & $0.85$ & $29.44$\\ 
AFM1-EWA-CV & $12.14$ & $9.46$ & $1.28$ & $0.96$ & $1.21$ & $0.89$ & $25.80$\\ 
\cmidrule{1-8}
\multicolumn{8}{l}{Panel C: Universe Size, $N = 1000$} \\\cmidrule{1-8}
EW-TQ & $14.63$ & $20.87$ & $0.70$ & $0.55$ & $0.69$ & $0.54$ & $58.56$\\ 
SC & $7.55$ & $11.88$ & $0.64$ & $0.38$ & $0.30$ & $0.04$ & $34.37$\\ 
EWA-SC & $6.84$ & $14.59$ & $0.47$ & $0.26$ & $0.10$ & $-0.10$ & $38.35$\\ 
LS & $10.06$ & $8.53$ & $1.18$ & $0.82$ & $1.01$ & $0.65$ & $30.96$\\ 
NL & $12.01$ & $8.00$ & $1.50$ & $1.12$ & $1.41$ & $1.03$ & $25.29$\\ 
EWA-CV & $12.70$ & $7.80$ & $\textbf{1.63}^{***}$ & $1.24$ & $1.55$ & $1.16$ & $23.19$\\ 
AFM1-NL & $11.96$ & $7.99$ & $1.50$ & $1.11$ & $1.40$ & $1.02$ & $25.37$\\ 
AFM1-EWA-CV & $12.72$ & $7.93$ & $1.60$ & $1.22$ & $1.52$ & $1.14$ & $24.82$\\ 
\bottomrule \end{tabular}
    \label{tab:mvoperf}
\end{table}
\vspace{\fill}

\newpage
\vspace*{\fill}
\begin{table}[H]
\caption{Weight Statistics}
    \captionsetup{font=footnotesize}
    \caption*{This table shows the average monthly weight statistics of Markowitz portfolios of various covariance estimators. The columns display the turnover (TO), the gross exposure (GE), the proportional leverage (PL), the minimum weight (MIN), the maximum weight (MAX), the standard deviation of weights (SD), and the mean absolute deviation from an equally-weighted portfolio (MAD). Panels A-C display the results for investment universe sizes $N \in \{ 100, 500, 1000 \}$. The summary statistics are obtained from an average over the 400 portfolio formation periods.}
    \centering
    \small
\centering
\small
\begin{tabular}{lccccccc} \toprule
 & TO & GE & PL & MIN & MAX & SD & MAD\\ \cmidrule{1-8}
\multicolumn{8}{l}{Panel A: Universe Size, $N = 100$} \\\cmidrule{1-8}
EW-TQ & $0.533$ & $1.000$ & $0.000$ & $0.000$ & $0.050$ & $0.020$ & $0.016$\\ 
SC & $1.231$ & $3.465$ & $0.444$ & $-0.101$ & $0.242$ & $0.050$ & $0.035$\\ 
EWA-SC & $1.542$ & $3.774$ & $0.448$ & $-0.119$ & $0.236$ & $0.053$ & $0.038$\\ 
LS & $1.162$ & $3.289$ & $0.434$ & $-0.093$ & $0.198$ & $0.045$ & $0.033$\\ 
NL & $1.126$ & $3.200$ & $0.432$ & $-0.089$ & $0.207$ & $0.044$ & $0.032$\\ 
EWA-CV & $1.250$ & $3.155$ & $0.426$ & $-0.090$ & $0.172$ & $0.042$ & $0.031$\\ 
AFM1-NL & $1.127$ & $3.211$ & $0.433$ & $-0.090$ & $0.210$ & $0.045$ & $0.032$\\ 
AFM1-EWA-CV & $1.198$ & $3.184$ & $0.427$ & $-0.086$ & $0.182$ & $0.043$ & $0.032$\\ 
\cmidrule{1-8}
\multicolumn{8}{l}{Panel B: Universe Size, $N =500$} \\\cmidrule{1-8}
EW-TQ & $0.509$ & $1.000$ & $0.000$ & $0.000$ & $0.010$ & $0.004$ & $0.003$\\ 
SC & $2.380$ & $6.037$ & $0.474$ & $-0.087$ & $0.253$ & $0.022$ & $0.012$\\ 
EWA-SC & $3.205$ & $7.237$ & $0.474$ & $-0.147$ & $0.275$ & $0.026$ & $0.014$\\ 
LS & $1.896$ & $5.134$ & $0.464$ & $-0.043$ & $0.137$ & $0.016$ & $0.010$\\ 
NL & $1.321$ & $3.953$ & $0.443$ & $-0.051$ & $0.196$ & $0.016$ & $0.008$\\ 
EWA-CV & $1.320$ & $3.745$ & $0.430$ & $-0.055$ & $0.187$ & $0.015$ & $0.007$\\ 
AFM1-NL & $1.328$ & $3.977$ & $0.444$ & $-0.049$ & $0.198$ & $0.016$ & $0.008$\\ 
AFM1-EWA-CV & $1.303$ & $3.782$ & $0.433$ & $-0.053$ & $0.190$ & $0.015$ & $0.007$\\ 
\cmidrule{1-8}
\multicolumn{8}{l}{Panel C: Universe Size, $N = 1000$} \\\cmidrule{1-8}
EW-TQ & $0.494$ & $1.000$ & $0.000$ & $0.000$ & $0.005$ & $0.002$ & $0.002$\\ 
SC & $7.178$ & $11.847$ & $0.484$ & $-0.122$ & $0.217$ & $0.019$ & $0.012$\\ 
EWA-SC & $9.467$ & $14.992$ & $0.485$ & $-0.194$ & $0.237$ & $0.024$ & $0.015$\\ 
LS & $2.663$ & $6.347$ & $0.467$ & $-0.027$ & $0.075$ & $0.009$ & $0.006$\\ 
NL & $1.430$ & $3.978$ & $0.431$ & $-0.038$ & $0.123$ & $0.008$ & $0.004$\\ 
EWA-CV & $1.268$ & $3.616$ & $0.416$ & $-0.025$ & $0.086$ & $0.006$ & $0.004$\\ 
AFM1-NL & $1.439$ & $3.996$ & $0.432$ & $-0.037$ & $0.124$ & $0.008$ & $0.004$\\ 
AFM1-EWA-CV & $1.265$ & $3.660$ & $0.418$ & $-0.026$ & $0.087$ & $0.006$ & $0.004$\\ 
\bottomrule \end{tabular}
    \label{tab:mvoweight}
\end{table}
\vspace{\fill}

\newpage

\vspace*{\fill}
\begin{table}[H]
\caption{Subsample Analysis}
    \captionsetup{font=footnotesize}
    \caption*{This table shows the annualized out-of-sample  standard deviation and information ratio of Markowitz portfolios of various covariance estimators over different subsamples. The columns display the subsample periods divided equally into Sample 1 (1986 through 1994), Sample 2 (1994 through 2003), Sample 3 (2003 through 2011), and Sample 4 (2011 through 2019). Panels A-C display the results for investment universe sizes $N \in \{ 100, 500, 1000 \}$. The daily out-of-sample returns are from $09/04/1986$ up to $12/31/2019$.  The most competitive value in the IR column is highlighted in \textbf{bold}.}
\centering
\small
\begin{tabular}{lccccccccccc} \toprule
& \multicolumn{2}{c}{Sample 1} &  & \multicolumn{2}{c}{Sample 2} &  & \multicolumn{2}{c}{Sample 3} &  & \multicolumn{2}{c}{Sample 4} \\ \cline{2-3} \cline{5-6} \cline{8-9} \cline{11-12}& SD & IR && SD & IR && SD & IR && SD & IR\\ \cmidrule{1-12}
\multicolumn{12}{l}{Panel A: Universe Size, $N = 100$} \\\cmidrule{1-12}
EW-TQ & $19.46$ & $0.79$&  & $28.18$ & $\textbf{0.90}$&  & $23.43$ & $0.28$&  & $16.33$ & $0.91$\\ 
SC & $14.00$ & $0.82$&  & $18.28$ & $0.58$&  & $13.19$ & $1.01$&  & $11.58$ & $1.14$\\ 
EWA-SC & $13.42$ & $\textbf{0.91}$&  & $17.76$ & $0.49$&  & $12.94$ & $0.92$&  & $11.08$ & $1.00$\\ 
LS & $13.94$ & $0.82$&  & $18.24$ & $0.60$&  & $13.11$ & $1.03$&  & $11.51$ & $1.18$\\ 
NL & $14.02$ & $0.83$&  & $18.17$ & $0.60$&  & $13.15$ & $1.03$&  & $11.48$ & $\textbf{1.19}$\\ 
EWA-CV & $13.69$ & $0.87$&  & $17.42$ & $0.61$&  & $12.60$ & $1.04$&  & $10.87$ & $1.13$\\ 
AFM1-NL & $14.03$ & $0.83$&  & $18.16$ & $0.61$&  & $13.17$ & $1.03$&  & $11.49$ & $1.18$\\ 
AFM1-EWA-CV & $13.82$ & $0.86$&  & $18.11$ & $0.58$&  & $12.68$ & $\textbf{1.09}$&  & $11.23$ & $1.09$\\ 
\cmidrule{1-12}
\multicolumn{12}{l}{Panel B: Universe Size, $N =500$} \\\cmidrule{1-12}
EW-TQ & $17.88$ & $0.88$&  & $25.14$ & $0.80$&  & $25.28$ & $0.39$&  & $15.78$ & $0.83$\\ 
SC & $10.38$ & $1.31$&  & $11.46$ & $1.01$&  & $11.69$ & $0.93$&  & $7.32$ & $1.23$\\ 
EWA-SC & $10.50$ & $1.12$&  & $12.61$ & $0.90$&  & $12.22$ & $0.83$&  & $7.42$ & $0.89$\\ 
LS & $9.93$ & $1.36$&  & $11.04$ & $1.07$&  & $11.21$ & $1.01$&  & $7.32$ & $\textbf{1.46}$\\ 
NL & $9.71$ & $1.38$&  & $10.82$ & $1.08$&  & $10.72$ & $1.21$&  & $6.95$ & $1.43$\\ 
EWA-CV & $9.11$ & $\textbf{1.44}$&  & $10.63$ & $\textbf{1.21}$&  & $10.13$ & $1.35$&  & $6.56$ & $1.36$\\ 
AFM1-NL & $9.69$ & $1.37$&  & $10.80$ & $1.08$&  & $10.72$ & $1.21$&  & $6.95$ & $1.41$\\ 
AFM1-EWA-CV & $9.40$ & $1.42$&  & $10.98$ & $1.14$&  & $10.24$ & $\textbf{1.36}$&  & $6.63$ & $1.33$\\ 
\cmidrule{1-12}
\multicolumn{12}{l}{Panel C: Universe Size, $N = 1000$} \\\cmidrule{1-12}
EW-TQ & $17.06$ & $0.95$&  & $23.81$ & $0.81$&  & $24.99$ & $0.40$&  & $16.15$ & $0.80$\\ 
SC & $11.45$ & $0.71$&  & $12.35$ & $0.82$&  & $13.79$ & $0.52$&  & $9.53$ & $0.51$\\ 
EWA-SC & $13.37$ & $0.39$&  & $17.00$ & $0.46$&  & $15.93$ & $0.57$&  & $11.41$ & $0.45$\\ 
LS & $8.06$ & $1.33$&  & $8.57$ & $1.49$&  & $9.98$ & $0.74$&  & $7.29$ & $1.28$\\ 
NL & $7.54$ & $1.66$&  & $8.43$ & $1.59$&  & $9.35$ & $1.26$&  & $6.38$ & $1.62$\\ 
EWA-CV & $7.28$ & $\textbf{1.69}$&  & $8.36$ & $\textbf{1.71}$&  & $8.95$ & $1.41$&  & $6.34$ & $\textbf{1.83}$\\ 
AFM1-NL & $7.53$ & $1.66$&  & $8.41$ & $1.58$&  & $9.35$ & $1.26$&  & $6.38$ & $1.60$\\ 
AFM1-EWA-CV & $7.46$ & $1.68$&  & $8.54$ & $1.62$&  & $9.06$ & $\textbf{1.42}$&  & $6.38$ & $1.83$\\ 
\bottomrule \end{tabular}
    \label{tab:mvoSubsample}
\end{table}
\vspace{\fill}

\newpage

\vspace*{\fill}
\begin{table}[H]
\caption{Sensitivity Analysis}
    \captionsetup{font=footnotesize}
    \caption*{This table shows the annualized out-of-sample standard deviation (in percentages) and information ratio of Markowitz portfolios estimated with EWA-SC and EWA-CV for different decay rates across different subsamples. The columns display the full sample period and the subsample periods divided equally into Sample 1 (1986 through 1994), Sample 2 (1994 through 2003), Sample 3 (2003 through 2011), and Sample 4 (2011 through 2019). Panels A-B and C-D display the results for investment universe sizes $N=100$ and $N=1000$, respectively.  The daily out-of-sample returns are from $09/04/1986$ up to $12/31/2019$. }
\centering
\small
\begin{tabular}{lcccccccccccccc} \toprule
& \multicolumn{2}{c}{Full Sample} &  &\multicolumn{2}{c}{Sample 1} &  & \multicolumn{2}{c}{Sample 2} &  & \multicolumn{2}{c}{Sample 3} &  & \multicolumn{2}{c}{Sample 4} \\ \cline{2-3} \cline{5-6} \cline{8-9} \cline{11-12} \cline{14-15}& SD & IR && SD & IR && SD & IR && SD & IR && SD & IR\\ \cmidrule{1-15}
\multicolumn{15}{l}{Panel A: EWA-SC, Universe Size: $N=100$} \\\cmidrule{1-15}
$\beta =0.999$ & $14.16$ & $0.83$&  & $13.63$ & $0.86$&  & $17.91$ & $0.55$&  & $12.96$ & $1.00$&  & $11.31$ & $1.09$\\ 
$\beta =0.997$ & $14.01$ & $0.78$&  & $13.42$ & $0.91$&  & $17.76$ & $0.49$&  & $12.94$ & $0.92$&  & $11.08$ & $1.00$\\ 
$\beta =0.995$ & $14.27$ & $0.74$&  & $13.67$ & $0.92$&  & $18.03$ & $0.42$&  & $13.30$ & $0.84$&  & $11.23$ & $0.96$\\ 
$\beta =0.992$ & $14.93$ & $0.68$&  & $14.30$ & $0.91$&  & $18.79$ & $0.32$&  & $14.00$ & $0.75$&  & $11.75$ & $0.91$\\ 
$\beta =0.990$ & $15.41$ & $0.64$&  & $14.77$ & $0.90$&  & $19.40$ & $0.25$&  & $14.46$ & $0.71$&  & $12.12$ & $0.87$\\ 
$\beta =0.985$ & $16.68$ & $0.55$&  & $15.99$ & $0.86$&  & $21.07$ & $0.11$&  & $15.53$ & $0.65$&  & $13.12$ & $0.80$\\ 
\cmidrule{1-15}
\multicolumn{15}{l}{Panel B: EWA-CV, Universe Size: $N=100$} \\\cmidrule{1-15}
$\beta =0.999$ & $14.10$ & $0.87$&  & $13.73$ & $0.87$&  & $17.77$ & $0.60$&  & $12.86$ & $1.06$&  & $11.18$ & $1.16$\\ 
$\beta =0.997$ & $13.85$ & $0.87$&  & $13.69$ & $0.87$&  & $17.42$ & $0.61$&  & $12.60$ & $1.04$&  & $10.87$ & $1.13$\\ 
$\beta =0.995$ & $13.90$ & $0.87$&  & $13.99$ & $0.86$&  & $17.36$ & $0.64$&  & $12.59$ & $1.02$&  & $10.83$ & $1.16$\\ 
$\beta =0.992$ & $14.05$ & $0.89$&  & $14.39$ & $0.87$&  & $17.43$ & $0.67$&  & $12.72$ & $1.01$&  & $10.84$ & $1.22$\\ 
$\beta =0.990$ & $14.15$ & $0.90$&  & $14.59$ & $0.85$&  & $17.54$ & $0.69$&  & $12.77$ & $1.01$&  & $10.86$ & $1.26$\\ 
$\beta =0.985$ & $14.31$ & $0.91$&  & $14.84$ & $0.81$&  & $17.70$ & $0.75$&  & $12.92$ & $1.00$&  & $10.93$ & $1.30$\\ 
\cmidrule{1-15}
\multicolumn{15}{l}{Panel C: EWA-SC, Universe Size: $N=1000$} \\\cmidrule{1-15}
$\beta =0.999$ & $12.32$ & $0.59$&  & $11.76$ & $0.63$&  & $13.16$ & $0.71$&  & $14.16$ & $0.52$&  & $9.79$ & $0.50$\\ 
$\beta =0.997$ & $14.59$ & $0.47$&  & $13.37$ & $0.39$&  & $17.00$ & $0.46$&  & $15.93$ & $0.57$&  & $11.41$ & $0.45$\\ 
$\beta =0.995$ & $17.80$ & $0.38$&  & $15.84$ & $0.18$&  & $21.93$ & $0.31$&  & $18.42$ & $0.64$&  & $13.99$ & $0.40$\\ 
$\beta =0.992$ & $21.39$ & $0.37$&  & $19.02$ & $0.10$&  & $26.83$ & $0.27$&  & $21.40$ & $0.73$&  & $17.05$ & $0.41$\\ 
$\beta =0.990$ & $23.24$ & $0.39$&  & $20.80$ & $0.11$&  & $29.22$ & $0.29$&  & $22.99$ & $0.77$&  & $18.58$ & $0.44$\\ 
$\beta =0.985$ & $27.44$ & $0.44$&  & $24.71$ & $0.16$&  & $34.75$ & $0.33$&  & $26.65$ & $0.80$&  & $22.00$ & $0.51$\\ 
\cmidrule{1-15}
\multicolumn{15}{l}{Panel D: EWA-CV, Universe Size: $N=1000$} \\\cmidrule{1-15}
$\beta =0.999$ & $7.89$ & $1.60$&  & $7.35$ & $1.71$&  & $8.37$ & $1.65$&  & $9.16$ & $1.36$&  & $6.43$ & $1.82$\\ 
$\beta =0.997$ & $7.80$ & $1.63$&  & $7.28$ & $1.69$&  & $8.36$ & $1.71$&  & $8.95$ & $1.41$&  & $6.34$ & $1.83$\\ 
$\beta =0.995$ & $7.88$ & $1.63$&  & $7.41$ & $1.62$&  & $8.50$ & $1.69$&  & $8.96$ & $1.48$&  & $6.42$ & $1.82$\\ 
$\beta =0.992$ & $8.10$ & $1.59$&  & $7.76$ & $1.52$&  & $8.74$ & $1.64$&  & $9.07$ & $1.52$&  & $6.64$ & $1.76$\\ 
$\beta =0.990$ & $8.25$ & $1.57$&  & $7.99$ & $1.46$&  & $8.90$ & $1.61$&  & $9.16$ & $1.52$&  & $6.76$ & $1.74$\\ 
$\beta =0.985$ & $8.61$ & $1.52$&  & $8.50$ & $1.34$&  & $9.30$ & $1.56$&  & $9.40$ & $1.49$&  & $7.02$ & $1.75$\\ 
\bottomrule \end{tabular}
    \label{tab:mvosensitive}
\end{table}
\vspace{\fill}

\newpage

\vspace*{\fill}
\begin{table}[H]
\caption{No Short-Sale Constraint}
    \captionsetup{font=footnotesize}
    \caption*{This table shows the annualized out-of-sample performances (in percentages) of Markowitz portfolios of various covariance estimators.  Short sales are prohibited due to a lower bound of zero on all portfolio weights. The columns display the average returns (AV), the standard deviation of returns (SD), the information ratio (IR), the Sharpe ratio (SR), and the maximum drawdown (MDD). The information ratio and Sharpe ratio net of transaction costs of 5 basis points are also reported and denoted as $\widetilde{\mathrm{IR}}$ and $\widetilde{\mathrm{SR}}$, respectively. Panels A-C display the results for investment universe sizes $N \in \{ 100, 500, 1000 \}$. The daily out-of-sample returns are from $09/04/1986$ up to $12/31/2019$. The most competitive value in the IR column is highlighted in \textbf{bold}. For the EWA-SC and EWA-CV portfolios, an asterisk indicates a significant outperformance of one portfolio versus the other in terms of IR: $^{***}$ denotes significance at the 0.01 level; $^{**}$ denotes significance at the 0.05 level; and $^{*}$ denotes significance at the 0.1 level.}
\centering
\small
\begin{tabular}{lccccccc} \toprule
 & AV & SD & IR & SR & $\widetilde{\mathrm{IR}}$ & $\widetilde{\mathrm{SR}}$ & MDD\\ \cmidrule{1-8}
\multicolumn{8}{l}{Panel A: Universe Size, $N = 100$} \\\cmidrule{1-8}
EW-TQ & $15.62$ & $22.30$ & $0.70$ & $0.56$ & $0.69$ & $0.55$ & $56.71$\\ 
SC & $13.44$ & $18.33$ & $0.73$ & $0.57$ & $0.71$ & $0.55$ & $48.35$\\ 
EWA-SC & $12.93$ & $17.82$ & $0.73$ & $0.55$ & $0.70$ & $0.53$ & $47.02$\\ 
LS & $13.44$ & $18.32$ & $0.73$ & $0.57$ & $0.71$ & $0.55$ & $47.94$\\ 
NL & $13.37$ & $18.33$ & $0.73$ & $0.56$ & $0.71$ & $0.54$ & $48.22$\\ 
EWA-CV & $13.01$ & $17.85$ & $0.73$ & $0.56$ & $0.71$ & $0.54$ & $47.62$\\ 
AFM1-NL & $13.39$ & $18.33$ & $0.73$ & $0.56$ & $0.71$ & $0.54$ & $48.23$\\ 
AFM1-EWA-CV & $13.48$ & $18.19$ & $\textbf{0.74}$ & $0.57$ & $0.72$ & $0.55$ & $46.30$\\ 
\cmidrule{1-8}
\multicolumn{8}{l}{Panel B: Universe Size, $N =500$} \\\cmidrule{1-8}
EW-TQ & $14.69$ & $21.44$ & $0.69$ & $0.54$ & $0.67$ & $0.53$ & $58.63$\\ 
SC & $13.29$ & $15.02$ & $0.88$ & $0.68$ & $0.86$ & $0.66$ & $38.68$\\ 
EWA-SC & $14.26$ & $14.28$ & $1.00$ & $0.79$ & $0.97$ & $0.76$ & $40.11$\\ 
LS & $13.34$ & $14.99$ & $0.89$ & $0.69$ & $0.87$ & $0.66$ & $38.61$\\ 
NL & $13.59$ & $14.94$ & $0.91$ & $0.71$ & $0.89$ & $0.68$ & $39.26$\\ 
EWA-CV & $14.38$ & $14.20$ & $\textbf{1.01}$ & $0.80$ & $0.99$ & $0.77$ & $37.75$\\ 
AFM1-NL & $13.58$ & $14.94$ & $0.91$ & $0.70$ & $0.89$ & $0.68$ & $39.36$\\ 
AFM1-EWA-CV & $13.95$ & $14.78$ & $0.94$ & $0.74$ & $0.92$ & $0.71$ & $36.40$\\ 
\cmidrule{1-8}
\multicolumn{8}{l}{Panel C: Universe Size, $N = 1000$} \\\cmidrule{1-8}
EW-TQ & $14.63$ & $20.87$ & $0.70$ & $0.55$ & $0.69$ & $0.54$ & $58.56$\\ 
SC & $12.60$ & $13.25$ & $0.95$ & $0.72$ & $0.92$ & $0.69$ & $33.25$\\ 
EWA-SC & $12.49$ & $12.56$ & $0.99$ & $0.75$ & $0.96$ & $0.72$ & $35.00$\\ 
LS & $12.76$ & $13.25$ & $0.96$ & $0.73$ & $0.94$ & $0.71$ & $34.29$\\ 
NL & $13.24$ & $13.37$ & $0.99$ & $0.76$ & $0.97$ & $0.74$ & $35.68$\\ 
EWA-CV & $12.75$ & $12.69$ & $\textbf{1.01}$ & $0.76$ & $0.98$ & $0.74$ & $36.83$\\ 
AFM1-NL & $13.24$ & $13.38$ & $0.99$ & $0.76$ & $0.97$ & $0.74$ & $35.73$\\ 
AFM1-EWA-CV & $12.96$ & $13.24$ & $0.98$ & $0.75$ & $0.95$ & $0.72$ & $36.48$\\ 
\bottomrule \end{tabular}
    \label{tab:mvonss}
\end{table}
\vspace{\fill}

\newpage

\vspace*{\fill}
\begin{table}[H]
\caption{Devolatization}
    \captionsetup{font=footnotesize}
    \caption*{This table shows the annualized out-of-sample performances (in percentages) of Markowitz portfolios for various estimators. The columns display the average returns (AV), the standard deviation of returns (SD), the information ratio (IR), the Sharpe ratio (SR), and the maximum drawdown (MDD). The information ratio and Sharpe ratio net of transaction costs of 5 basis points are also reported and denoted as $\widetilde{\mathrm{IR}}$ and $\widetilde{\mathrm{SR}}$, respectively. Panels A-C display the results for investment universe sizes $N \in \{ 100, 500, 1000 \}$. The daily out-of-sample returns are from $09/04/1986$ up to $12/31/2019$. The most competitive value in the IR column is highlighted in \textbf{bold}. }
\centering
\small
\begin{tabular}{lccccccc} \toprule
 & AV & SD & IR & SR & $\widetilde{\mathrm{IR}}$ & $\widetilde{\mathrm{SR}}$ & MDD\\ \cmidrule{1-8}
\multicolumn{8}{l}{Panel A: Universe Size, $N = 100$} \\\cmidrule{1-8}
VOL & $13.85$ & $18.79$ & $0.74$ & $0.57$ & $0.71$ & $0.55$ & $50.92$\\ 
DCC-NL & $12.41$ & $14.51$ & $0.86$ & $0.65$ & $0.74$ & $0.53$ & $37.29$\\ 
EWA-CV & $13.62$ & $14.45$ & $\textbf{0.94}$ & $0.73$ & $0.85$ & $0.64$ & $44.01$\\ 
AFM1-DCC-NL & $12.16$ & $14.16$ & $0.86$ & $0.64$ & $0.79$ & $0.57$ & $34.91$\\ 
AFM1-EWA-CV & $12.45$ & $14.34$ & $0.87$ & $0.66$ & $0.79$ & $0.58$ & $31.73$\\ 
\cmidrule{1-8}
\multicolumn{8}{l}{Panel B: Universe Size, $N =500$} \\\cmidrule{1-8}
VOL & $13.68$ & $16.00$ & $0.85$ & $0.66$ & $0.82$ & $0.63$ & $42.52$\\ 
DCC-NL & $11.63$ & $9.07$ & $1.28$ & $0.95$ & $1.12$ & $0.78$ & $24.36$\\ 
EWA-CV & $11.90$ & $8.82$ & $1.35$ & $1.00$ & $1.21$ & $0.87$ & $29.30$\\ 
AFM1-DCC-NL & $12.18$ & $8.83$ & $1.38$ & $1.03$ & $1.24$ & $0.89$ & $28.00$\\ 
AFM1-EWA-CV & $12.42$ & $8.72$ & $\textbf{1.42}$ & $1.08$ & $1.29$ & $0.94$ & $26.16$\\  
\cmidrule{1-8}
\multicolumn{8}{l}{Panel C: Universe Size, $N = 1000$} \\\cmidrule{1-8}
VOL & $13.28$ & $14.74$ & $0.90$ & $0.69$ & $0.87$ & $0.66$ & $43.16$\\ 
DCC-NL & $10.37$ & $7.19$ & $1.44$ & $1.02$ & $1.26$ & $0.83$ & $25.86$\\ 
EWA-CV & $10.64$ & $7.09$ & $1.50$ & $1.07$ & $1.35$ & $0.92$ & $30.29$\\ 
AFM1-DCC-NL & $11.09$ & $6.88$ & $1.61$ & $1.17$ & $1.44$ & $1.00$ & $21.41$\\ 
AFM1-EWA-CV & $11.15$ & $6.85$ & $\textbf{1.63}$ & $1.18$ & $1.46$ & $1.01$ & $23.39$\\
\bottomrule \end{tabular}
    \label{tab:mvodevol}
\end{table}
\vspace{\fill}

\end{document}